\documentclass[authoryear,preprint,review,12pt]{elsarticle}
\usepackage[left=2.5 cm,top=3.0cm,bottom=3.0cm,right=2.5 cm]{geometry}

\usepackage{amssymb}
\usepackage{amsfonts,amsmath,bm} 
\usepackage{color}
\usepackage[colorlinks=true, allcolors=blue]{hyperref}

\usepackage{lineno}

\journal{Computer Methods in Applied Mechanics and Engineering}

\begin{document}

\begin{frontmatter}



\title{One-shot learning for the complex dynamical behaviors of weakly nonlinear 
forced oscillators}

\author[label1,label2]{Teng Ma}
\author[label2]{Luca Rosafalco\corref{cor1}}
\author[label1]{Wei Cui}
\author[label1,label3]{Lin Zhao}
\author[label2]{Attilio Frangi}
\affiliation[label1]{organization={State Key Lab of Disaster Reduction in Civil Engineering},
            addressline={Tongji University},
            city={Shanghai},
            postcode={200092},
            state={Shanghai},
            country={P. R. China}}

\affiliation[label2]{organization={Department of Civil and Environmental Engineering},
            addressline={Politecnico di Milano},
            city={Milan},
            postcode={20133},
            state={Lombardia},
            country={Italy}}

\affiliation[label3]{organization={Guangxi Laboratory of Whole Life Safety for Land-Sea Corridor Engineering},
            addressline={Guangxi University},
            city={Nanning},
            postcode={530004},
            state={Guangxi},
            country={P. R. China}}

\cortext[cor1]{email: luca.rosafalco@polimi.it}


\begin{abstract}
Extrapolative prediction of complex nonlinear dynamics remains a central challenge in engineering. This study proposes a one-shot learning method to identify global frequency-response curves from a single excitation time history by learning governing equations. We introduce MEv-SINDy (Multi-frequency Evolutionary Sparse Identification of Nonlinear Dynamics) to infer the governing equations of non-autonomous and multi-frequency systems. The methodology leverages the Generalized Harmonic Balance (GHB) method to decompose complex forced responses into a set of slow-varying evolution equations. We validated the capabilities of MEv-SINDy on two critical Micro-Electro-Mechanical Systems (MEMS). These applications include a nonlinear beam resonator and a MEMS micromirror. Our results show that the model trained on a single point accurately predicts softening/hardening effects and jump phenomena across a wide range of excitation levels. This approach significantly reduces the data acquisition burden for the characterization and design of nonlinear microsystems.

\end{abstract}

\begin{graphicalabstract}
\begin{figure}[h]
    \centering
    \includegraphics[width=1\linewidth]{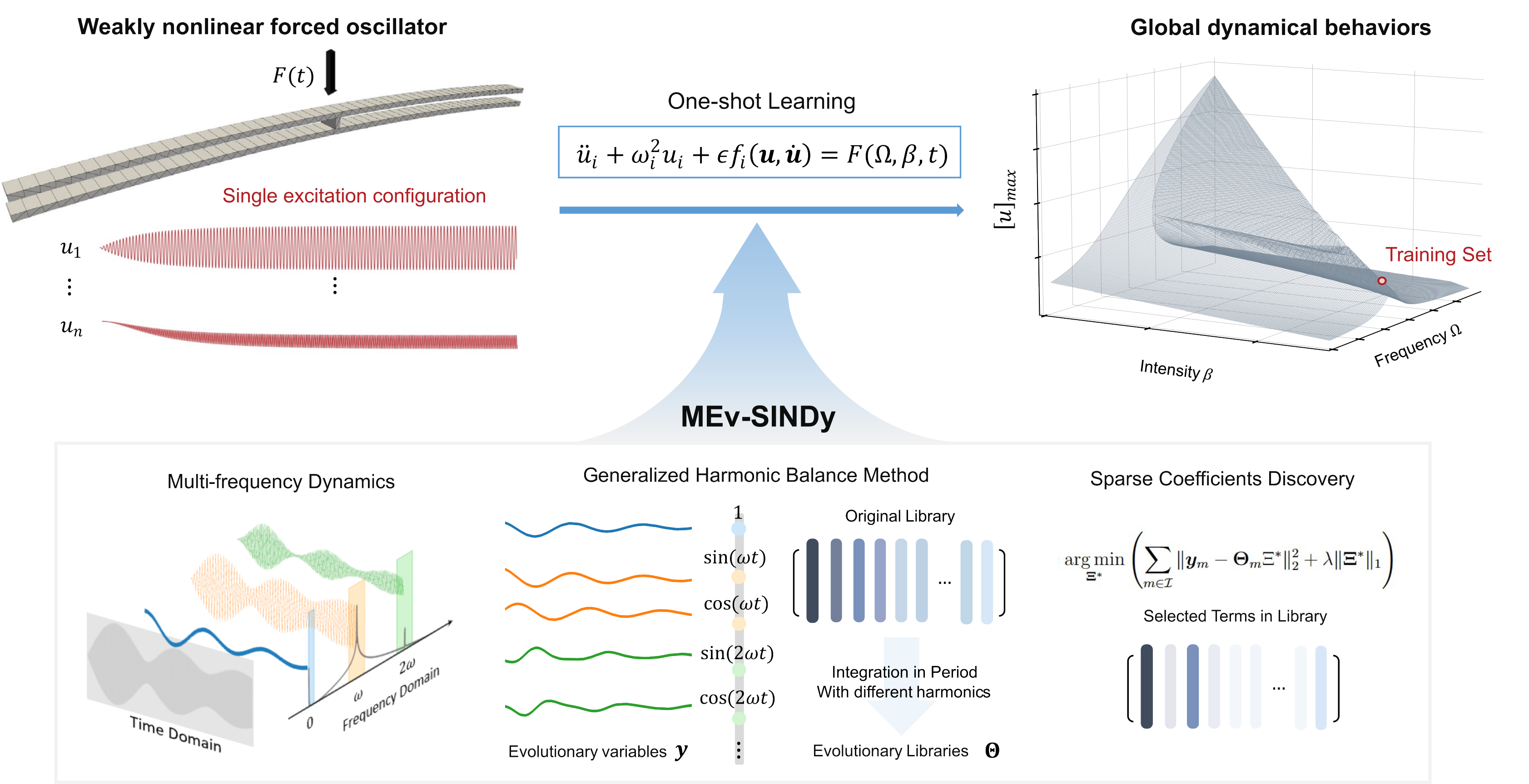}
    \label{fig:placeholder}
\end{figure}

\end{graphicalabstract}

\begin{highlights}
\item A one-shot learning method identifies global nonlinear frequency-response curves of weakly nonlinear forced oscillators from a single excitation time history by inferring the governing equations.
\item The equation learning framework for weakly nonlinear systems is upgraded from autonomous single-frequency dynamics to non-autonomous multi-frequency dynamics. This advancement is achieved by incorporating the generalized harmonic balance theory.
\end{highlights}


\end{frontmatter}



\section{Introduction}
\label{sec1}

The rapid advancement of computational resources has enabled the high-fidelity simulation of complex physical phenomena, even within intricate multi-physics and multi-scale frameworks. However, characterizing the complex dynamical behaviors of physical systems remains a formidable challenge. In practical engineering applications, such as MEMS resonators and sensors, understanding the nonlinear response—including frequency shifts, hardening/softening effects, and jump phenomena—requires extensive frequency and amplitude sweeps. When relying on full-order models (FOMs), such as the finite element method (FEM), each individual simulation in the time domain or parametric space can be computationally very demanding. Consequently, exploring a broad range of initial conditions and parameter combinations becomes a computationally prohibitive task or even unfeasible for real-time monitoring and design optimization. Unlike simple output estimation, capturing the evolution of a high-dimensional nonlinear field is intrinsically difficult due to its time-dependent nature and the sensitivity of nonlinear regimes. These challenges drive the urgent need for efficient yet accurate reduced-order models (ROMs) that can bypass the heavy burden of FOMs while preserving the essential physics of the dynamical system.

Efficient ROM construction begins with dimensionality reduction~\citep{van2009dimensionality}. 
This process projects high-dimensional FOM states onto a low-dimensional manifold from data previously collected from full-order simulations or experiments~\citep{huang2017dimensionality}. Historically, the reduced basis method has been the standard for building these reduced spaces~\citep{maday2002reduced,quarteroni2015reduced,benner2017model}. It often utilizes Proper Orthogonal Decomposition (POD)~\citep{amsallem2012nonlinear,pagani2018numerical,gobat2022reduced} to approximate solutions. However, POD-based techniques are typically intrusive. They require direct access to the internal operators of the FOM. This requirement makes it difficult to use them with proprietary or commercial simulation software. Machine learning has emerged as a powerful non-intrusive alternative for ROM construction. These methods reduce dimensionality directly from data streams. They do not require access to the FOM operators. Dynamic Mode Decomposition (DMD)~\citep{schmid2011applications,le2017higher} is a prominent linear framework in this category. It extracts dominant frequencies and spatial modes from snapshot data. However, DMD is limited by its linear assumption. Autoencoder (AE) networks~\citep{zhai2018autoencoder,romor2023non,gonzalez2018deep} provide a nonlinear alternative. They are particularly effective at uncovering latent features in complex systems. These networks offer better nonlinear compression than linear methods like POD or DMD. Advanced frameworks like deep learning-based ROMs (DL-ROMs)~\citep{fresca2022deep,franco2023deep} and their enhanced version through POD (POD-DL-ROMs)~\citep{fresca2022pod} have further improved this process. They perform dimensionality reduction and learn the parameter-to-solution map at the same time.

Data-driven ROMs face significant challenges when predicting scenarios outside the training set. These methods are essentially interpolation tools. They provide high accuracy only within the observed parameter ranges. When these models encounter unseen parameter values, their performance often degrades. This lack of generalization limits their use in exploring unknown physical regimes. Furthermore, constructing reliable non-intrusive ROMs requires a large volume of training data. Generating these datasets through FOM simulations is computationally expensive. This high cost reduces the overall efficiency of the modeling process. There is a clear need for a framework that offers better extrapolation. Such a framework should also require less training data to maintain physical consistency.

Equation discovery offers a powerful alternative to black-box models. This approach infers governing equations directly from data. There are two main methodological paradigms in this field. The first paradigm is symbolic discovery~\citep{schmidt2009distilling, ruan2025discovering}. It seeks to recover closed-form representations of physical laws. The second paradigm is sparse regression~\citep{brunton2016discovering, gao2022autonomous}. This method identifies parsimonious models from a predefined library of candidate functions. These techniques were first developed for ordinary differential equations. Researchers later extended them to partial differential equations~\citep{rudy2017data,rao2023encoding}. Recent advances in machine learning have improved these methods. They are now robust against sparse or noisy data~\citep{chen2021physics,hirsh2022sparsifying,niven2024dynamical,fung2025rapid}. This makes them viable for real-world scientific discovery. Equation discovery has catalyzed the field of scientific machine learning. It helps uncover interpretable laws in cyber-physical systems~\citep{yuan2019data}, fluid dynamics~\citep{ma2023data,loiseau2018sparse}, biological and chemical networks~\citep{mangan2016inferring,hoffmann2019reactive} and complex networks~\citep{gao2024learning,yu2025discover}. The main appeal of this approach is the generation of an explicit ROM. The resulting model takes the form of a system of ordinary differential equations. Sparsity constraints ensure that the latent dynamics remain interpretable. Users can then analyze the identified system using standard numerical tools for time integration.

Most physical and engineering systems exhibit weakly nonlinear oscillatory behaviors with weak linear dissipation. These include applications in structural engineering and micro-electro-mechanical systems (MEMS). Weak nonlinearities often arise from geometric effects. They can lead to pronounced impacts on dynamical performance. However, existing methods often overlook these challenges. They focus on extracting only the most dominant governing equations. This process often discards terms that seem insignificant but are physically important. Recent research has introduced evolutionary-based learning to address this issue. A prominent example is EvLOWN~\citep{ma2026encoding}, which stands for Encoding Cumulation to Learn Perturbative Nonlinear Oscillatory Dynamics. This method reformulates optimization objectives using the method of averaging. It bridges the gap between weak nonlinearities and slow-varying evolutionary variables. Only the terms most informative about amplitude and phase evolutions are selected. However, EvLOWN has specific limitations. It can only identify autonomous single-frequency responses but cannot handle multi-frequency oscillations or non-autonomous dynamics. In this work, we propose a novel data-driven approach. We call it Multi-frequency Evolutionary Sparse Identification of Nonlinear Dynamical systems (MEv-SINDy). This framework is designed to learn the governing equations of multi-frequency forced weakly nonlinear oscillators. It extends the advantages of evolutionary learning to more complex forced and multi-frequency regimes.

The capability to identify governing equations from multi-frequency data opens the door to a more ambitious paradigm. This paradigm is known as one-shot learning~\citep{1597116}. It involves training models to recognize global patterns based on only a single data point. The concept originated in computer vision but remains rare in scientific machine learning (SciML)~\citep{Ivanov2020PhysicsbasedPN,DARCY2023133583,jiao2025oneshot}. Most data-driven ROMs still require massive datasets to characterize physical systems~\citep{conti2023reduced,gobat2022reduced,franco2023deep}. However, generating such snapshots for complex nonlinear oscillators is computationally expensive. We leverage MEv-SINDy to achieve one-shot learning for forced nonlinear dynamics. In our framework, the identified governing equations represent the ``general knowledge'' of the system. We extract this knowledge from a single excitation configuration. This analytical representation allows the model to predict dynamical behaviors under entirely different parameters. Consequently, MEv-SINDy can predict the complete frequency-response curve from one observation. This strategy eliminates the need for exhaustive training data. It also provides a robust tool for exploring unknown physical regimes with minimal cost. To our knowledge, this is one of the first applications of one-shot learning for multi-frequency forced oscillators.

\begin{figure}[h]
    \centering
    \includegraphics[width=1\linewidth]{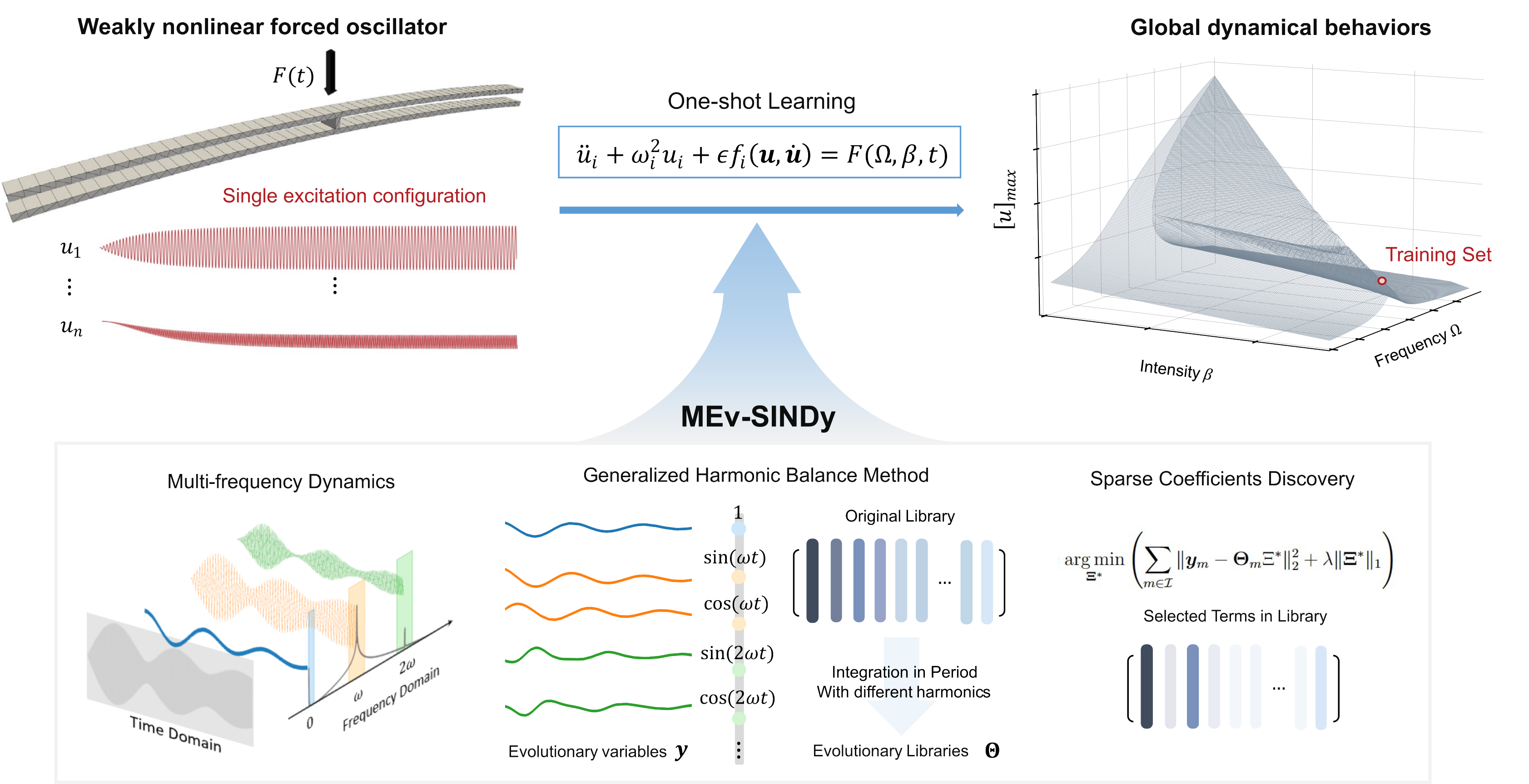}
    \caption{\textbf{One-Shot Learning framework of weakly nonlinear forced oscillator}: The starting point of our framework is the dynamical response at a single excitation configuration, which is also the training data of MEv-SINDy. During the training stage, the multi-frequency dynamics are separated into distinct harmonic components. We use the generalized harmonic balance method to reformulate the sparse regression problem into an evolution space. A two-stage sparse regression process then discovers the final coefficient vector. The primary output of MEv-SINDy is the explicit governing equation of the weakly nonlinear forced oscillator. Finally, we apply a continuation method to the inferred equations to compute the global frequency-response curves.}
    \label{fig:abstract}
\end{figure}

Schematic diagram of this work is shown in Figure~\ref{fig:abstract}. The paper is structured as follows. Section 2 details the methodology of the proposed MEv-SINDy framework. This section introduces the Generalized Harmonic Balance (GHB) method as an upgrade to the averaging method. The evolutionary variables are expanded from simple amplitude and phase to general harmonic coefficients. Based on GHB, we reformulate the sparse regression optimization targets. We also describe a joint sparse regression method designed for multiple equations. Section 3 presents numerical validations using a 1-DOF oscillator with quadratic and cubic nonlinearities. These tests validate the effectiveness of each step in the MEv-SINDy process. Section 4 considers two high-dimensional applications. These include a Beam MEMS resonator and a MEMS micromirror. We demonstrate that our approach achieves one-shot learning for the frequency-response curve. To further analyze the robustness of the method, we examine the influence of training point selection in Section 4. We describe a multi-point training strategy to improve predictive performance. Finally, Section 5 provides concluding remarks and summarizes the findings.

\section{Methodology}

\subsection{Problem setup}

The starting point is to consider the general form of weakly nonlinear oscillators with external forcing:


\begin{equation}\label{eq:WNO_start}
    \ddot{x}_i(t)+\omega_i^2x_i(t)+\epsilon f_i(\bm{x}(t),\bm{\dot{x}}(t))=F_i(t),
\end{equation}
for $i=1,\ldots,k$,where: $\bm{x}(t)=(x_1(t),\cdots,x_k(t))\in \mathbb{R}^k$ is the displacement vector at time $t$; $\omega_i$ is the frequency of oscillator $i$; $f_i(\bm{x}(t),\bm{\dot{x}}(t))$  is a smooth function of the velocity $\bm{\dot{x}}$ and the displacement $\bm{x}$ representing the weak nonlinearity and weak dissipation, which are from geometric, internal, material, contact and coupling nonlinearities; $F_i(t)$ is the external actuation that depends on time $t$; $\epsilon$ is a small parameter quantifying the strength of nonlinearity compared to dominating linear spring effects $\omega_i^2 x_i$. Eq. \eqref{eq:WNO_start} is well suited for modeling various classes of systems, whose dimensionality $k$ represents the number of degrees of freedom arising, for example, from spatial discretization techniques (e.g.\ finite elements or finite volumes) applied to a system of partial differential equations (PDEs) governing the underlying physical problem. 

Due to the weak nonlinearity $f_i$, the steady-state periodic response of the system governed by Eq. \eqref{eq:WNO_start} under different harmonic actuation $F_i$ exhibits distinct characteristics in terms of amplitude and phase, collectively referred to as the frequency-response curves (FRCs), which characterize the dynamic properties of the system. Typically, the actuation can be parameterized by two features, frequency and intensity, both of which strongly affect the steady-state periodic response. The goal of this study is to utilize data snapshots from a single time history and a given actuation to accurately predict responses under any other actuation conditions, thus reconstructing all the possible FRCs of the system. The proposed methodology employs only few set of time history of displacement $\bm{x}$ and its time-derivative $\dot{\bm{x}}$. In general, $\dot{\bm{x}}$ can be either computed directly or approximated numerically from $\bm{x}$. 

\subsection{Sparse Identification of Nonlinear Dynamics: SINDy}

To identify the system in Eq. \eqref{eq:WNO_start}, it is possible to rely on a linear combination of a set of predetermined functions collected in a library, as done in the SINDy method,~\citep{brunton2016discovering}. To apply SINDy, snapshots of $\bm{x}$ at different time instants are collected in the following matrix:
\begin{equation}
     \mathbf{X} = 
    \begin{bmatrix}
        x_1(t_1) & x_2(t_1) & \cdots & x_n(t_1) \\
        x_1(t_2) & x_2(t_2) & \cdots & x_n(t_2) \\
        \vdots & \vdots & \ddots & \vdots \\
        x_1(t_T) & x_2(t_T) & \cdots & x_n(t_T)
    \end{bmatrix}.
    \label{eq:snapshotMatrix}
\end{equation}
where: $x_i(t_j)$ is the i-th entry of $\mathbf{x}$ at the $j$-th time instant; $T$ is the number of time instants for each snapshot. Similarly, a matrix $\dot{\mathbf{X}}$ is constructed by collecting $\dot{\bm{x}}$. 

A library $\boldsymbol{\Theta}(\mathbf{x})=\left[\theta_1(\mathbf{x}), \ldots, \theta_p(\mathbf{x})\right]\in\mathbb{R}^p$ of $p$ candidate functions is selected. In Sec. \ref{sec:OMP}, it will be discussed how to retain only the most relevant functions among this set of candidates. The matrix $\boldsymbol{\Theta}(\mathbf{X})\in\mathbb{R}^{T\times p}$ is thus constructed by applying $\boldsymbol{\Theta}$ to the rows of $\mathbf{X}$, for example, as in the following:
\begin{equation}
\boldsymbol{\Theta}(\mathbf{X})=\begin{bmatrix}
        | & | & | & | & & | & | & \\
        1 & \mathbf{X} & \mathbf{X}^{P_2} & \mathbf{X}^{P_3} & \cdots & \text{sin}(\mathbf{X}) & \text{cos}(\mathbf{X}) & \cdots \\
        | & | & | & | & & | & | &
    \end{bmatrix}.
\label{eq:functionLabels}
\end{equation}

Any nonlinear function, such as polynomial and trigonometric, cna be included in the library.

As mentioned, the model to be identified is a linear combination of the candidate functions, with combination coefficients stored in $\boldsymbol{\Xi}=\left[ \boldsymbol{\xi}_1, \ldots, \boldsymbol{\xi}_k\right]$ with $\boldsymbol{\xi}_k\in \mathbb{R}^{p}$. Combination coefficients are determined by solving a regression problem with a sparsity promoting term, as in the following:

\begin{equation}
\begin{aligned}
    \mathop{\arg\min}\limits_{\bm{\Xi}} \|\bm{\ddot{x}}- \bm{\Theta}(\bm{x},\bm{\dot{x}},t)\bm{\Xi}\|_2^2+\lambda\| \bm{\Xi}\|_1
\end{aligned}
\label{eq:L1norm}
\end{equation}

where $||\cdot||_2^2$ is the least squares term; $||\cdot||_1$ is the $L_1$ norm of the coefficient vector $\bm{\Xi}$, which is the sum of absolute values of the coefficients; $\lambda$ is the regularization parameter that controls the strength of the penalty. In Sec. \ref{sec:OMP}, Eq. \eqref{eq:L1norm} will be modified for a better coupling with the proposed algorithm.

\subsection{Weakly nonlinear oscillator inference: EvLOWN}

It is worth noting that most existing model identification frameworks such SINDy do not target the specific difficulties introduced by weak nonlinearities in practical systems. Indeed, these methods identify parsimonious governing equations by neglecting library terms weighted by very small terms. While this approach is meaningful in general, it may become a major drawback in the context of WNOs, where the very necessary components for accurately capturing the dynamics may be discarded as the dynamics is governed by the linear term $\omega^2_ix_i$.

To address this gap, \cite{ma2026encoding} proposed a data-driven approach called EvLOWN (Evolutionary Learning Oscillator with Weak Nonlinearity) for discovering weakly nonlinear oscillator. EvLOWN reformulates the original system of ODEs into an evolutionary-variable representation via averaging theory, naturally separating terms by their order of magnitude. The goal of EvLOWN is to discover the autonomous oscillatory dynamical system:
\begin{equation}\label{eq:WNO_withoutForcing}
    \ddot{x}_i(t)+\omega_i^2x_i(t)+\epsilon f_i(\bm{x}(t),\bm{\dot{x}}(t))=0
\end{equation}
by learning the evolution of amplitude and phase, rather than directly learning the state variables $\bm{x}$.

Leveraging averaging theory~\citep{sanders2007averaging}, also known as the Krylov-Bogoliubov-Mitropolsky method~\citep{volosov1962averaging}, we approximate the response of the oscillator \eqref{eq:WNO_withoutForcing} when the nonlinear effect is weak (i.e., $\epsilon$ is small) as:
\begin{equation}\label{eq:WNO_solution0}
    x_i(t) = A_i(t)\sin(\hat{\omega}_i t + \phi_i(t))
\end{equation}
where: $\hat{\omega}_i$ is the oscillating frequency; $A_i$ and $\phi_i$ are the amplitude and phase, respectively.

In WNOs, the amplitude $A_i(t)$ and phase $\phi_i(t)$ evolve much more slowly over time compared to the oscillatory term, changing little over a single period $2\pi/\hat{\omega}_i$ \cite{nayfeh2008applied}. This separation of timescales allows us to average the system dynamics over one oscillation cycle, yielding explicit relationships between the weak nonlinearities and the time evolution of the amplitude $A_i(t)$ and phase $\phi_i(t)$:
\begin{equation}
\begin{aligned}
    A_i'(t) & =\frac{A_i(t+\pi/\hat{\omega}_i)-A_i(t-\pi/\hat{\omega}_i)}{2\pi/\hat{\omega}_i}=\frac{\epsilon_i}{2\pi}\int_{t-\pi/\hat{\omega}_i}^{t+\pi/\hat{\omega}_i}f_i(\bm{x}(s), \dot{\bm{x}}(s))\cos(\hat{\omega}_i s+\phi_i(s))ds,  \\
    \phi_i'(t) & = \frac{\phi_i(t+\pi/\hat{\omega}_i)-\phi_i(t-\pi/\hat{\omega}_i)}{2\pi/\hat{\omega}_i} = -\frac{\epsilon_i}{2A_i(t)\pi}\int_{t-\pi/\hat{\omega}_i}^{t+\pi/\hat{\omega}_i}f_i(\bm{x}(s), \dot{\bm{x}}(s))\sin(\hat{\omega}_i s+\phi_i(s))ds,
\end{aligned}
\label{eq:MoA method}
\end{equation}
where $(\cdot)'$ denotes the rate of change over one period.

To identify the first-order ODEs governing the evolution of amplitude and phase, we integrate the basis terms in the original library $\bm{\Theta}$ over a single period, which quantifies their contributions to the averaged dynamics. This transformation results in two new libraries: $\bm{\Theta}_A$ for amplitude evolution and $\bm{\Theta}_\phi$ for phase evolution. The discovery task now becomes identifying a sparse subset of basis terms from these libraries that best approximate $A_i'(t)$ and $\phi_i'(t)$. Critically, this averaging process naturally separates terms by order of magnitude, allowing weak nonlinear contributions to be isolated from dominant dynamics.

\subsection{Generalized Harmonic Balance method}
\label{sec:method-hb}


As the magnitude of external forcing increases in the forced oscillator Eq.~ \eqref{eq:WNO_start}, the contribution of nonlinearities in the system, such as quadratic and cubic terms, becomes significant. 
Consequently, the assumption of a single-frequency solution, see Eq.~\eqref{eq:WNO_solution0} becomes inadequate for complex forced dynamics. To address this issue, we employ the generalized harmonic balance method to replace the averaging method, thus extending the EvLOWN approach to account for multiple-frequency responses. The solution of the weakly nonlinear forcing oscillator can be approximated as follows~\citep{luo2012approximate}:
\begin{equation}\label{eq:WNO_solution0}
    x^*_i(t) = a^{(0)}_i(t)+\sum_{n=1}^{N}\{b_i^{(n)}(t)\cos(n\hat{\omega}_i t)+c_i^{(n)}(t)\sin(n\hat{\omega}_i t)\}
\end{equation}
where the superscript $(\cdot)^{(n)}$ indicates the order of the harmonic frequency, corresponding to the $n$-multiple of the fundamental frequency $\hat{\omega}_i$. The terms $a_i^{(0)}(t)$, $b_i^{(n)}(t)$, and $c_i^{(n)}(t)$ are the slowly varying parameters associated with each harmonic component $n\hat{\omega}_i$.

The first- and second-order time derivatives of $\bm{x}^*(t)$ are:
\begin{equation}\label{eq:WNO_derivate1}
    \dot{x}_i^*(t) = \dot{a}_i^{(0)}(t)+\sum_{n=1}^{N}\{[\dot{b}_i^{(n)}(t)+n\hat{\omega}_i c_i^{(n)}(t)]\cos(n\hat{\omega}_i t)+[\dot{c}_i^{(n)}-n\hat{\omega}_i b_i^{(n)}(t)]\sin(n\hat{\omega}_i t)\}
\end{equation}
\begin{equation}\label{eq:WNO_derivate2}
\begin{aligned}
    \ddot{x}_i^*(t) = \ddot{a}_i^{(0)}(t)+\sum_{n=1}^{N}\{[\ddot{b}_i^{(n)}(t)+2n\hat{\omega}_i \dot{c}_i^{(n)}(t)-(n\hat{\omega}_i)^2 b_i^{(n)}(t)]\cos(n\hat{\omega}_i t)\\+[\ddot{c}_i^{(n)}(t)-2n\hat{\omega}_i \dot{b}_i^{(n)}(t)-(n\hat{\omega}_i)^2c_i^{(n)}(t)]\sin(n\hat{\omega}_i t)\}
\end{aligned}
\end{equation}

Supposing that $a_i^{(0)}(t)$, $b_i^{(n)}(t)$, and $c_i^{(n)}(t)$ vary slowly with time, we substitute Eqs. \eqref{eq:WNO_solution0}-\eqref{eq:WNO_derivate2} into Eq. \eqref{eq:WNO_start}, and then we average for each harmonic terms of $\cos(n\omega_i t)$ and $\sin(n\omega_i t)$ obtaining for $n=1,2,\cdots,N$:

\begin{equation}
\begin{aligned}
    \ddot{a}_i^{(0)}(t)+\omega_i^2a_0(t)=-&\frac{\hat{\omega}_i}{2\pi}\int_{t-\pi/\hat{\omega}_i}^{t+\pi/\hat{\omega}_i}[f_i^*(\bm{x},\bm{\dot{x}},s)]ds\\
    \ddot{b}_i^{(n)}(t)+2\hat{\omega}_i n\dot{c}_i^{(n)}(t)-\hat{\omega}_i^2(n^2-1)b_i^{(n)}(t)=-&\frac{\hat{\omega}_i}{\pi}\int_{t-\pi/\hat{\omega}_i}^{t+\pi/\hat{\omega}_i}[f_i^*(\bm{x},\bm{\dot{x}},s)]\cos(n\hat{\omega}_is)ds\\
    \ddot{c}_i^{(n)}(t)-2\hat{\omega}_i n\dot{b}_i^{(n)}(t)-\hat{\omega}_i^2(n^2-1) c_i^{(n)}(t)=-&\frac{\hat{\omega}_i}{\pi}\int_{t-\pi/\hat{\omega}_i}^{t+\pi/\hat{\omega}_i}[f_i^*(\bm{x},\bm{\dot{x}},s)]\sin(n\hat{\omega}_is)ds,
\end{aligned}
\label{eq:GHB method}
\end{equation}
where:
\begin{equation}
    f_i^*(\bm{x},\dot{\bm{x}},s) = \epsilon f_i(\bm{x}^*(s),\dot{\bm{x}}^*(s))-F_i(\bm{x}^*(s),\dot{\bm{x}}^*(s),s)
\end{equation}

It is worth stressing that the only assumption is related to the slow temporal variation of $a_i^{(0)}(t)$, $b_i^{(n)}(t)$, and $c_i^{(n)}(t)$, while no limitations on the type of response (steady state or transient) is instead required. In the next section, we will show how to exploit the evolutionary formulation in Eq.~\eqref{eq:GHB method} to identify the system described, in its original formulation, by Eq.~\eqref{eq:WNO_start}.

\subsection{Multiple slow-varying evolutions evaluations}
\label{sec:method-filter}

We now describe how to extract the slowly varying evolutionary variables $a_i^{(0)}(t)$, $b_i^{(n)}(t)$, and $c_i^{(n)}(t)\ (n = 1, 2, \ldots)$ from the observed signal $x(t)$. These variables are subsequently used to perform sparse regression on the evolutionary formulation reported in Eq.~\eqref{eq:GHB method} in order to identify the unknown components of the governing equations, namely $\omega_i$, $f_i^*$.

Given an input signal $\bm{x}(t) = [x_1(t), \ldots, x_d(t)]^\mathsf{T}$, the oscillation frequency in the $i$-th dimension can be directly computed from the Fourier transform $\mathcal{F}(x_i(t)):\mathbb{R}\rightarrow\mathbb{R}$, with:

\begin{equation}
\label{eq:fft}
    \hat{\omega}_i = \mathop{\arg\max}\limits_{\hat{\omega_i}} \mathcal{F}(x_i(t)) = \mathop{\arg\max}\limits_{\hat{\omega_i}} \int_{-\infty}^{+\infty} x_i(t)e^{-i\hat{\omega}_i t}dt 
\end{equation}

It is worth noting that the oscillation circular frequency $\hat{\omega}_i$ differs from the natural circular frequency $\omega_i$. This deviation arises from the presence of weakly nonlinear terms and external forcing. To illustrate this, we consider a simple example of a linearly damped oscillator described by $\ddot{x} + 2\beta\dot{x} + \omega^2x = F\cos(\Omega t)$. Under the assumption of light damping ($\beta \ll 1$), the solution of the linear damped oscillator is given by: 
\[
x(t)=x_0e^{-\beta t}\cos(\sqrt{\omega^2-\beta^2}t+\phi)+\frac{f}{\sqrt{(\omega^2-\Omega^2)^2+4\beta^2\omega^2}}\cos(\Omega t-\varphi)
\]

Therefore, the oscillation circular frequency $\hat{\omega}$ obtained from Eq.~\eqref{eq:fft} lies within the range $(\sqrt{\omega^2 - \beta^2}, \Omega)$, and the value of $\hat{\omega}$ depends on the selected portion of the data $x(t)$. If the transient response dominates, $\hat{\omega}$ will tend to
$\sqrt{\omega^2 - \beta^2}$; otherwise, it will be close to $\Omega$.

\begin{figure}[h!]
\centering
\includegraphics[width = \textwidth]{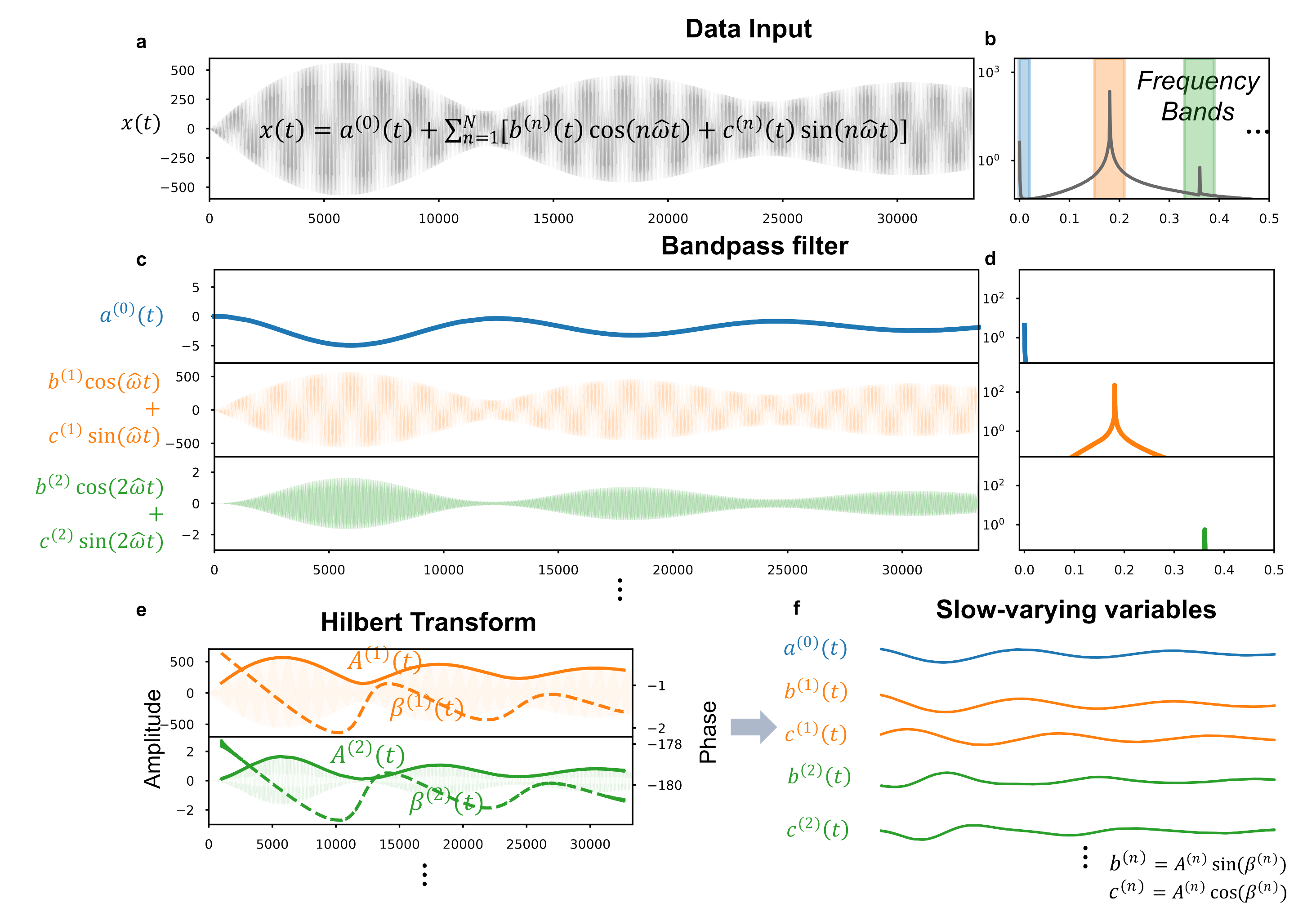}
\caption{
Identification process of the slowly varying evolutionary variables $a^{(0)}(t)$, $b^{(n)}(t)$, and $c^{(n)}(t)$ by combining a bandpass filter and the Hilbert transform.
(a) Input signal $x(t)$, representing the time-domain response of a weakly nonlinear forced oscillator;
(b) $x(t)$ consists of multiple narrowband harmonic components centered at $n\hat{\omega}$, as shown in the frequency domain. Several frequency bands are defined to isolate each $n$-th harmonic component;
(c) Time-domain representation and (d) corresponding frequency-domain representation of the filtered narrowband signals;
(e) Hilbert transform applied to each $n$-th harmonic component to extract the instantaneous amplitude $A^{(n)}(t)$ and phase $\beta^{(n)}(t)$;
(f) The slowly varying coefficients $b^{(n)}(t)$ and $c^{(n)}(t)$ are calculated from the extracted amplitude and phase, while the zero-order term directly corresponds to $a^{(0)}(t)$.
}
\label{fig:multiplevariables}
\end{figure}

To obtain the multiple slow-varying evolutionary variables $a_i^{(0)}(t)$, $b_i^{(n)}(t)$, and $c_i^{(n)}(t)\ (n = 1, 2, \ldots)$ from the given time history $x_i(t)$, we proposed a novel identification method involving the Hilbert transform. Unlike what was proposed in~\citep{ma2026encoding}, the Hilbert transform cannot be directly applied to extract the slowly varying amplitude and phase, because the dynamic response of a weakly nonlinear forced oscillator contains multiple frequency components. To address this, we define $N$ frequency bands to address the limitations of standard Fourier analysis. In classical spectral analysis, isolated peaks represent constant coefficients. However, our evolutionary coefficients depend on time. This time-dependency smears the frequency peaks in the spectrum. In particular, we determine the value of $N$ depends on the number of distinct frequency peaks observed in the frequency spectrum. When, as in Figure~\ref{fig:multiplevariables}a, $x(t)$ consists of the zero, first and second harmonic components, we set $N = 3$. Each frequency band to be centered at $n\hat{\omega}$ and has a bandwidth of $\hat{\omega}/4$ (Figure~\ref{fig:multiplevariables}b). By passing the signal through the corresponding bandpass filter, each harmonic component is isolated, as shown in Figure~\ref{fig:multiplevariables}c. In practice, Finite Impulse Response (FIR) filters are designed using the window method. This procedure is implemented in readily available packages such as \texttt{scipy.signal}.

After having obtained the slow-varying evolutionary variables $a_i^{(0)}(t)$, $b_i^{(n)}(t)$, and $c_i^{(n)}(t)$ through the Hilbert transform, we can elaborate Eq. \eqref{eq:WNO_solution0} as:
\begin{equation}
\label{eq:superposition}
    x_i(t) = \sum_{i=0}^Nx^{(n)}_i(t),
\end{equation}
with:
\begin{equation}
\begin{aligned}
    x_i^{(n)}(t)=\left\{
    \begin{array}{ll}
        a_i^{(0)}(t)   & n=0 \\
        b_i^{(n)}(t)\cos(n\hat{\omega}t)+c_i^{(n)}(t)\sin(n\hat{\omega}t)  & n=1,2,\cdots
    \end{array}
    \right.
\end{aligned}
\label{eq:harmonic comonent}
\end{equation}

The zero-order harmonic component directly corresponds to $a^{(0)}(t)$. While for $n>0$, the slow-varying evolutionary variables are calculated  from amplitude and phase. The Hilbert transform $\mathcal{H}$ is used to compute the analytic signal $\hat{x}_i^{(n)}(t) = x_i^{(n)}(t) + i\mathcal{H}(x_i^{(n)}(t))$. The evolutionary variables $A^{(n)}_i(t)$ and $\beta^{(n)}_i(t)$ of time $t$ can be considered as the averaging value of instantaneous evolutionary variables in one period:

\begin{equation}
\begin{aligned}
    A_i^{(n)}(t) &= \frac{\int_{t-\pi/\hat{\omega}_i}^{t+\pi/\hat{\omega}_i} |\hat{x}_i^{(n)}(s)| ds}{2\pi/\hat{\omega}_i}  \\
    \beta_i^{(n)}(t) &= \frac{\int_{t-\pi/\hat{\omega}_i}^{t+\pi/\hat{\omega}_i} {\rm Arg} (\hat{x}_i^{(n)}(s)) ds}{2\pi/\hat{\omega}_i}
\end{aligned}
\label{eq:hilbert}
\end{equation}

As Figure~\ref{fig:multiplevariables}f shows, the evolutionary variable $b^{(n)}_i(t)$ and $c_i^{(n)}(t)$ of higher harmonic components can be calculated as Eq.~\eqref{eq:hilbert2}:

\begin{equation}
\begin{aligned}
    b^{(n)}_i(t)=A_i^{(n)}(t)\sin(\beta_i^{(n)}(t))\\
    c^{(n)}_i(t)=A_i^{(n)}(t)\cos(\beta_i^{(n)}(t))\\
\end{aligned}
\label{eq:hilbert2}
\end{equation}

\subsection{Reformulation of the Regression Problem in the Evolutionary Domain}
\label{sec:method-reformulation}

In order to accurately identify the governing dynamics of weakly nonlinear forced oscillators, it is essential to reformulate the regression problem in a form that enhances the visibility of nonlinear effects.

Sharing the same starting point of EvLOWN, we transform the problem into an evolutionary domain. In the proposed MEv-SINDy approach, the system behavior is represented by slowly varying evolutionary variables based on generalized harmonic balance method. This reformulation allows the nonlinear interactions to be more effectively captured and facilitates the application of sparse regression for discovering the underlying equations. We thus exploit SINDy to approximate the evolutionary formulation reported in Eq. \eqref{eq:GHB method}, instead of the original formulation (Eq.~\eqref{eq:WNO_start}). This is equivalent to perform regularized regression reported in Eq. \eqref{eq:L1norm}, but here involving the evolutionary instead of the original variables to identify the weakly nonlinear coefficients.

According to Sec.~\ref{sec:method-filter}, the slow-varying evolutionary variables $a_i^{(0)}(t)$,$b_i^{(n)}(t)$ and $c_i^{(n)}(t)\ (n = 1, 2, \ldots)$, together with the oscillation frequency $\hat{\omega}_i$, can be computed from the given time history $x_i(t)$. Eq.~\eqref{eq:GHB method} can then be reorganized such that known quantities are collected in the left hand side, while in the terms to be inferred are assembled in the right hand side.

\begin{equation}
\begin{aligned}
    \ddot{a}_i^{(0)}(t)+\hat{\omega}_i^2a_0(t)&=-\frac{\hat{\omega}_i}{2\pi}\int_{t-\pi/\hat{\omega}_i}^{t+\pi/\hat{\omega}_i}\hat{f}_i^*(\bm{x},\bm{\dot{x},s})ds\\
    \ddot{b}_i^{(n)}(t)+2\hat{\omega}_i n\dot{c}_i^{(n)}(t)-\hat{\omega}_i^2(n^2-1)b_i^{(n)}(t)&=-\frac{\hat{\omega}_i}{\pi}\int_{t-\pi/\hat{\omega}_i}^{t+\pi/\hat{\omega}_i}\hat{f}_i^*(\bm{x},\bm{\dot{x},s})\cos(n\hat{\omega}_is)ds\\
    \ddot{c}_i^{(n)}(t)-2\hat{\omega}_i n\dot{b}_i^{(n)}(t)-\hat{\omega}_i^2(n^2-1) c_i^{(n)}(t)&=-\frac{\hat{\omega}_i}{\pi}\int_{t-\pi/\hat{\omega}_i}^{t+\pi/\hat{\omega}_i}\hat{f}_i^*(\bm{x},\bm{\dot{x},s})\sin(n\hat{\omega}_is)ds\\
    \hat{f}_i^*(\bm{x},\dot{\bm{x}},s) &= (\omega_i^2-\hat{\omega}_i^2)x_i^*(s)+ \epsilon f_i(\bm{x}^*(s),\dot{\bm{x}}^*(s))-F_i(\bm{x}^*(s),\dot{\bm{x}}^*(s),s)\\
    n&=1,2,\cdots,N
\end{aligned}
\label{eq:GHB method_rewritten}
\end{equation}
where $\hat{f}_i^*(\bm{x},\dot{\bm{x}},t)=\bm{\Theta}^*(\bm{x},\dot{\bm{x}},t)\bm{\Xi}^*$ are unknown functions expressed as a sparse combination of a set of linear and nonlinear candidate basis function. Thanks to this rearrangement, the system is expressed in a standard regression form suitable for sparse identification. Unlike conventional sparse regression, our objective is to find a sparse coefficient vector $\bm{\Xi}^*$ that simultaneously satisfies all equations in Eq.~\eqref{eq:GHB method_rewritten}, rather than a single one. This gives rise to the concept of joint sparse regression, where the goal is to estimate a shared sparse representation across multiple regression targets. Mathematically, given $M = 2N + 1$ equations reported in the following:
\begin{equation}
    \bm{y}_i(t) = \bm{\Theta}_i(t) \bm{\Xi}^*,
\label{eq:multitargets}
\end{equation}
\begin{equation}
\bm{y}_i(t)=\begin{bmatrix}
y_i^{(0)}(t)\\ 
y_i^{(1,\mathrm{cos})}(t)\\   
y_i^{(1,\mathrm{sin})}(t)\\ 
y_i^{(2,\mathrm{cos})}(t)\\
y_i^{(2,\mathrm{sin})}(t)\\
\vdots
\end{bmatrix}=\begin{bmatrix}
\ddot{a}_i^{(0)}(t)+\hat{\omega}_i^2a_0(t)\\ 
\ddot{b}_i^{(1)}(t)+2\hat{\omega}_i \dot{c}_i^{(1)}(t)\\   
\ddot{c}_i^{(1)}(t)-2\hat{\omega}_i \dot{b}_i^{(1)}(t)\\ 
\ddot{b}_i^{(2)}(t)+4\hat{\omega}_i \dot{c}_i^{(2)}(t)-3\hat{\omega}_i^2b_i^{(2)}(t)\\
\ddot{c}_i^{(2)}(t)-4\hat{\omega}_i \dot{b}_i^{(2)}(t)-3\hat{\omega}_i^2c_i^{(2)}(t)\\
\vdots
\end{bmatrix},
\label{eq:multiy}
\end{equation}

\begin{equation}
\bm{\Theta}_i(t)=\begin{bmatrix}
\bm{\Theta}_i^{(0)}(t)\\ 
\bm{\Theta}_i^{(1,\mathrm{cos})}(t)\\   
\bm{\Theta}_i^{(1,\mathrm{sin})}(t)\\ 
\bm{\Theta}_i^{(2,\mathrm{cos})}(t)\\
\bm{\Theta}_i^{(2,\mathrm{sin})}(t)\\
\vdots
\end{bmatrix}=\begin{bmatrix}
-\frac{\hat{\omega}_i}{2\pi}\int_{t-\pi/\hat{\omega}_i}^{t+\pi/\hat{\omega}_i}\bm{\Theta}^*(s)ds\\ 
-\frac{\hat{\omega}_i}{\pi}\int_{t-\pi/\hat{\omega}_i}^{t+\pi/\hat{\omega}_i}\bm{\Theta}^*(s)\cos(\hat{\omega}_is)ds\\   
-\frac{\hat{\omega}_i}{\pi}\int_{t-\pi/\hat{\omega}_i}^{t+\pi/\hat{\omega}_i}\bm{\Theta}^*(s)\sin(\hat{\omega}_is)ds\\ 
-\frac{\hat{\omega}_i}{\pi}\int_{t-\pi/\hat{\omega}_i}^{t+\pi/\hat{\omega}_i}\bm{\Theta}^*(s)\cos(2\hat{\omega}_is)ds\\
-\frac{\hat{\omega}_i}{\pi}\int_{t-\pi/\hat{\omega}_i}^{t+\pi/\hat{\omega}_i}\bm{\Theta}^*(s)\sin(2\hat{\omega}_is)ds\\
\vdots
\end{bmatrix},
\label{eq:multitheta}
\end{equation}
the objective is to determine a common coefficient vector $\bm{\Xi}^*$ that minimizes the overall residual while maintaining sparsity. Only when this condition is met, the reconstructed system can accurately recover the original formulation expressed in Eq.~\eqref{eq:WNO_start}. In Eqs.~\eqref{eq:multiy} and \eqref{eq:multitheta}, the superscript $(\cdot)^{(n,t)}$ identifies the harmonic term in Eq.~\eqref{eq:GHB method_rewritten}. Specifically, $n \in \mathbb{R}^+$ represents the order of the harmonic frequency, corresponding to the $n$-fold multiple of the fundamental frequency $\hat{\omega}_i$, and $\{\mathrm{cos}, \mathrm{sin}\}$ indicates the cosine and sine components.

\subsection{Joint sparse regression across multiple equations}
\label{sec:OMP}

To infer the sparse coefficient vector $\bm{\Xi}^*$ that best satisfies all evolutionary equations, we propose a joint sparse regression method across multiple equations all evolutionary equations Eq.~\eqref{eq:multitargets}, rather than fitting each equation independently. The problem is formulated as minimizing the total residual across the $2N + 1$ evolutionary equations derived from the weakly nonlinear oscillator, each corresponding to a harmonic component over $N+1$ frequencies, with an $L_1$ regularization term to ensure sparsity.


A key algorithmic choice is setting which harmonics are considered in the evolutionary representation among $\bm{y}_{full}(t)= [y^{(0)}(t), y^{(1,cos)}(t), y^{(1,sin)}(t), y^{(2,cos)}(t), y^{(2,sin)}(t), \cdots]$. 
Once set, goal of the optimization is to identify a sparse coefficient vector $\bm{\Xi}^*$ that simultaneously satisfies all evolutionary equations derived from weakly nonlinear forcing oscillator as shown in the following:
\begin{equation}
    \bm{\Xi}^* = \mathop{\arg\min}\limits_{\bm{\Xi}^*}\left(\sum_{m\in\mathcal{I_S}} \|\bm{y}^{(m)}-\bm{\Theta}^{(m)}\Xi^*\|_2^2+\lambda\|\bm{\Xi}^*\|_1\right)
\label{eq:optimizationeffective}
\end{equation}
where:  $\mathcal{I_S}$ collects all the index of the effective harmonic components (e.g, $0,(1,\cos),(1,\sin),\cdots$); $\bm{y}^{(m)}$ and $\bm{\Theta}^{(m)}$ denote the response and library matrices for the $m$-th equation; $\lambda>0$ controls the sparsity level of the solution. We introduce a two-phase sparse regression method to solve Eq.~\eqref{eq:optimizationeffective}. The first phase utilizes a convex optimization algorithm to select an initial subset of the library. In this work, we employ the CVXOPT solver~\citep{vandenberghe2010cvxopt} to handle the $L_1$ regularization term. This step provides a preliminary sparse structure for the coefficient vector. The second phase involves a sparse fine-tuning process that removes terms with negligible contributions, thus ensuring a concise and generalized model. We apply an approach inspired by orthogonal matching pursuit to narrow down the model space. We select only the most relevant basis functions from the initial subset. The contribution $c_{ij}$ of each basis function $j$ for the $i$-th dimension is calculated using Eq.~\eqref{eq:contributions}:
\begin{align}
    c_{ij} = \int_0^T \bigg(\sum _{m\in\mathcal{I_S}}\frac{\left(\hat{\Theta}_{mij}(t)\Xi_{ij}^*\right)^2}{\left(y_{im}(t)\right)^2} \bigg) dt  \label{eq:contributions} 
\end{align}
These values are then normalized to a range between 0 and 1. Finally, we fine-tune the model by applying a cut-off threshold. Any basis function with a contribution value lower than this threshold is removed from the library. This two-phase strategy ensures the identification of a parsimonious dynamical system. It maintains high predictive accuracy while preventing overfitting to numerical noise.

\subsection{Parameter continuation method}
\label{sec:method-matcont}

In nonlinear dynamical systems, periodic solutions often depend on one or more system parameters, such as the excitation frequency and intensity 
and can be computed using several classical numerical techniques, such as the harmonic balance method~\citep{cheung1990application}, the shooting method~\citep{osborne1969shooting}, and collocation-based approaches~\citep{carrington2021using}. These methods transform the governing differential equations into nonlinear algebraic systems by enforcing periodicity conditions, and the steady-state responses are then obtained through iterative solvers such as the Newton–Raphson method~\citep{ypma1995historical}. However, these traditional approaches are often computationally expensive because they require repeated integrations or nonlinear solves for each excitation parameter. In addition, they usually depend on initial guesses that are close to stable solutions, which makes them incapable of accurately tracing unstable branches or capturing turning points in the frequency–response curves. As a result, their applicability is limited when a complete description of the system dynamics, including both stable and unstable periodic responses, is required.

The continuation method provides a systematic procedure to trace the evolution of these solutions as parameters vary, allowing for the identification of critical phenomena such as bifurcations or stability transitions. Consider a general system of nonlinear algebraic equations derived from the steady or periodic conditions of a dynamical system:

\begin{equation}
    \mathrm{G}(\bm{u},\zeta)=0
\label{eq:continuation}
\end{equation}
where $\bm{u}$ denotes the unknown state vector (e.g., displacement and velocity), and $\zeta$ represents the control parameter to be varied (e.g., frequency and amplitude of external forcing).

A straightforward approach would be to increment $\zeta$ in small steps and solve for $\bm{u}$ at each step using the previous solution as the initial guess $\zeta_{k+1}=\zeta_k+\Delta\zeta$. However, this simple ``parameter sweep'' method fails e.g.\ near turning points or folds in the solution curve, where the slope $\partial\bm{u}/\partial\zeta$ becomes infinite and the solution path cannot be followed uniquely.

To overcome this limitation, arc-length continuation (also known as pseudo–arc-length continuation) is commonly employed~\citep{krauskopf2007numerical}. In this method, the solution path is parameterized by an artificial variables, representing the arc length along the curve in the $(\bm{u},\zeta)$ space. The continuation step is then constrained by:

\begin{equation}
    (\Delta \bm{u})^T(\Delta \bm{u})+(\Delta \zeta)^2=(\Delta s)^2
\label{eq:continuationStep}
\end{equation}
where $\Delta s$ is the prescribed step size along the solution trajectory.

By solving the extended system composed of Eq.~\eqref{eq:continuation} together with the arc-length constraint, the continuation algorithm can robustly track both stable and unstable branches of the solution family, even across fold and bifurcation points. This technique provides a powerful and general framework for the numerical exploration of nonlinear response characteristics. These methods are implemented in many ready-to-use packages like \texttt{Auto07p} ~\citep{doedel1998auto}, that implements collocation methods in \texttt{FORTRAN} to perform numerical continuation and bifurcation analysis; \texttt{Manlab}, a \texttt{Matlab} tool that uses HB methods and Asymptotic Numerical Method~\citep{guillot2019taylor}; \texttt{Nlvib}, that also exploits HB methods~\citep{krack2019harmonic}, and many others among which we mention \texttt{COCO}~\citep{dankowicz2013recipes}. The numerical examples addressed in this work will be solved using the Matlab \texttt{MATCONT} package~\citep{dhooge2006matcont}, that exploits collocation methods to perform the continuation of periodic orbits.

\subsection{Mean chamfer distance of frequency response curves (MCDRC)}

Frequency response curves (FRCs) are the primary goals for one-shot learning and prediction, describing complex dynamic behaviors of weakly nonlinear oscillators. FRCs characterize a steady-state output quantity of interest, such as maximum deflection, across various load multipliers $\beta$ for all excitation frequency values $\Omega$ within a specified range. To quantitatively assess the predictive ability, an error metric named Mean Chamfer Distance of frequency response curves (MCDRC) between two FRCs is defined herein. 

It is worth mentioning that FRCs are generally not single-valued functions of the excitation frequency. For a given frequency, multiple periodic solutions may exist due to nonlinear effects such as jump phenomena, hysteresis, and multi-stability. Furthermore, continuation-based computations generally produce differently spaced samples along solution branches, and distinct response branches may be specified using varying frequency grids. These properties violate the assumptions required by most pointwise or function-based error measures.

To address this limitation, we represent FRCs as point sets in the frequency-response space. This allows to naturally support multi-valued responses, discontinuities, and nonuniform sampling, hence offering a consistent and stable foundation for quantitative comparison of nonlinear frequency response curves.

\begin{figure}[h]
    \centering
    \includegraphics[width=1\linewidth]{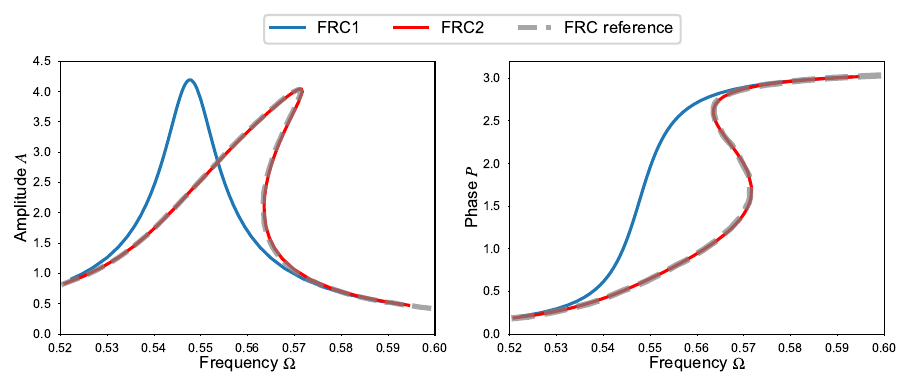}
    \caption{Two examples of mean chamfer distance of frequency response functions (MCDRC): FRC1 (blue solid line), whose MCDRC is 0.24, exhibits a noticeable deviation from the reference (gray dashed line); FRC2 (red solid line) has a MCDRC of 0.01 shows a much better correlation with the reference. 
    }
    \label{fig:MCDRF example}
\end{figure}

We then exploit the Chamfer distance $d_{ch}(\mathcal{C}\to \mathcal{C}_{ref})$~\citep{butt1998optimum} to establish a novel metric termed Mean Chamfer Distance of Frequency Response Curves (MCDRC). 
The chamfer distance $d_{ch}(\mathcal{C}\to \mathcal{C}_{ref})$ is defined as the average distance from each point on the predicted curve $\mathcal{C}=\{(x_i,y_i)\}_{i=1}^N$ to its nearest neighbor on the reference curve $\mathcal{C}_{ref}=\{(x_j,y_j)\}_{j=1}^M$. It can be computed as:
\begin{equation}
    d_{ch}(\mathcal{C}\to \mathcal{C}_{ref})=\frac{1}{N}\sum_{p\in \mathcal{C}}\min_{q\in \mathcal{C}_{ref}}\|p-q\|
\label{eq:optimizationeffective}
\end{equation}
By selecting the oscillation amplitude ($A$) and phase ($P$) as primary steady-state descriptors, the FRC is discretized into two point sets: $\mathcal{C}^A=\{(\Omega_i,A_i)\}_{i=1}^N$ and $\mathcal{C}^P=\{(\Omega_i,P_i)\}_{i=1}^N$. Here, $A$ and $P$ denote the amplitude and the phase lag relative to the external excitation, respectively. The MCDRC is defined as follows:
\begin{equation}
    \text{MCDRF} = d_{ch}(\mathcal{C}^A\to C^A_{ref}) + d_{ch}(\mathcal{C}^P\to \mathcal{C}^P_{ref})
\label{eq:optimizationeffective}
\end{equation}
Prior to calculating the distance between points, a max-min scaling normalization pre-processing step is introduced to ensure that the scaling disparity between the axes does not spoil the representativeness of MCDRC. Figure~\ref{fig:MCDRF example} presents an illustrative example of MCDRC comparing two predicted FRCs against a reference FRC. The MCDRC for FRC1 is 0.24, indicating poor predictive accuracy, whereas the value for FRC2 is 0.01, representing excellent agreement with the reference. This comparison provides an intuitive demonstration of how MCDRC quantifies the similarity between FRCs.

\section{Numerical Validations}

To demonstrate the effectiveness of the proposed framework, we first consider a benchmark one-dimensional (1D) forced oscillator characterized by linear dissipation, weak quadratic and cubic nonlinearities. This model is widely used in nonlinear dynamics to represent various physical systems~\citep{kudryashov2021generalized}. The objective is to evaluate the framework capability in formulating the governing equations and to numerically verify the equivalence between the original formulation (Eq.~\eqref{eq:WNO_start}) and the generalized evolutionary formulation (Eq.~\eqref{eq:GHB method}). The equation of motion is expressed as:

\begin{equation}\label{eq:WNO_case1}
    \ddot{x}+\omega^2x+c\dot{x}+\alpha_1x^2+\alpha_2x^3=\beta\cos(\Omega t)
\end{equation}
where $x$,$\dot{x}$ and $\ddot{x}$ denote the displacement, velocity and acceleration of the oscillator, respectively; $\omega$ is the natural frequency, $\alpha_1,\alpha_2 \in \mathbb{R}$ are parameters characterizing the weak nonlinearities, $c\in \mathbb{R}$ is the linear damping coefficient, $\beta$ and $\Omega$ are the amplitude and frequency of the external forcing.  To generate the training trajectory, the oscillator described by Eq.~\eqref{eq:WNO_case1} is numerically integrated using the following model parameters: $c=1\times 10^{-2},\alpha_1=1\times10^{-2}, \alpha_2=1\times 10^{-4}, \omega=2, \Omega=1.999,
\beta=0.5$, with initial states $x_0=[0,0]$. 
Indeed, the magnitudes of $\alpha_1,\alpha_2$ and $c$ are orders of magnitude smaller than $\omega$.

\begin{figure}[h!]
\centering
\includegraphics[width = 11cm]{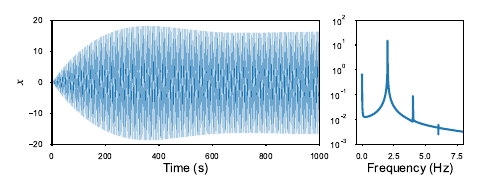}
\caption{
Benchmark 1D forced oscillator. Training data of numerical quadratic and cubic nonlinear oscillator in time and frequency domain.
}
\label{fig:numericaldata}
\end{figure}

The resulting time-domain and frequency-domain responses are illustrated in Figure~\ref{fig:numericaldata}. The frequency spectrum clearly indicates that the dominant dynamical response occurs near the fundamental frequency of approximately 2. Additionally, two secondary peaks are observed at the frequencies 0 and 4, which correspond to the DC component and second-order harmonic, respectively. These multi-frequency features underline the necessity of a higher-order analytical treatment. It is worth mentioning that the previously proposed EvLOWN approach~\citep{ma2026encoding} is unable to handle the presence of higher-order harmonics.

Through this numerical example, we aim to demonstrate the efficacy of the proposed method regarding three key aspects:

\begin{itemize}
   \item First, to justify the necessity of incorporating multi-frequency responses in forced nonlinear oscillators;
    \item Second, to validate the evaluation framework for multiple slowly varying evolutionary variables and the generalized harmonic balance (GHB) approach;
    \item Third, to showcase the effectiveness of the method in both governing equation inference and prediction of complex nonlinear dynamical behaviors.
\end{itemize}

\subsection{Multi-frequency response under external forcing}

The emergence of multi-frequency response stems from the system nonlinearities, specifically where the nonlinear restoring force exhibits a nonlinear dependence on the displacement $x$. As the amplitude of displacement increases, the contributions of higher-order nonlinear terms may grow order of times faster (e.g.\ $x^2$ and $x^3$), leading to the emergence of pronounced harmonic components in the response spectrum. For forced system, the external excitation may drive the system into a regime where higher-order nonlinearities dominate, resulting in a rich multi-frequency spectrum. Conversely, in the absence of external forcing, a free oscillator typically exhibits diminished vibration amplitudes, effectively oscillating at a single frequency (i.e., its natural frequency), as other harmonic components are negligible .

\begin{figure}[h!]
\centering
\includegraphics[width = 9cm]{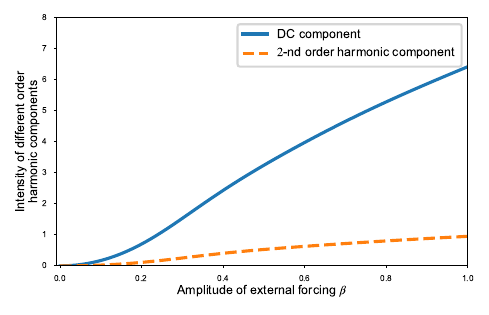}
\caption{
Benchmark 1D forced oscillator. Influence of external forcing amplitude $\beta$ on the intensities of the DC offset (zero-order) and second-order harmonic components.
}
\label{fig:extra_harmonic}
\end{figure}

To quantitatively illustrate this phenomenon, we perform a numerical test using the nonlinear oscillator ruled by Eq.~\eqref{eq:WNO_case1}. All parameters remain constant except for the amplitude of the external forcing $\beta$, which varies from 0 to 1. The additional harmonic components, specifically the DC offset (zero-order) and second-order harmonic responses, are evaluated under different values of $\beta$. The initial conditions are adjusted to $x_0=[1,0]$. As shown in Figure~\ref{fig:extra_harmonic}, when $\beta$ approaches zero, the magnitudes of these extra harmonic components vanish, indicating a nearly quasi-linear single-frequency response. However, as $\beta$ increases, these higher-order harmonics exhibit a steady growth. This trend confirms that intensified external excitation amplifies nonlinear effects, thereby underscoring the necessity of incorporating multi-frequency responses when analyzing forced nonlinear oscillators.

\subsection{Numerical validation of generalized harmonic balance}

\begin{figure}[h!]
    \centering
    \includegraphics[width=1\linewidth]{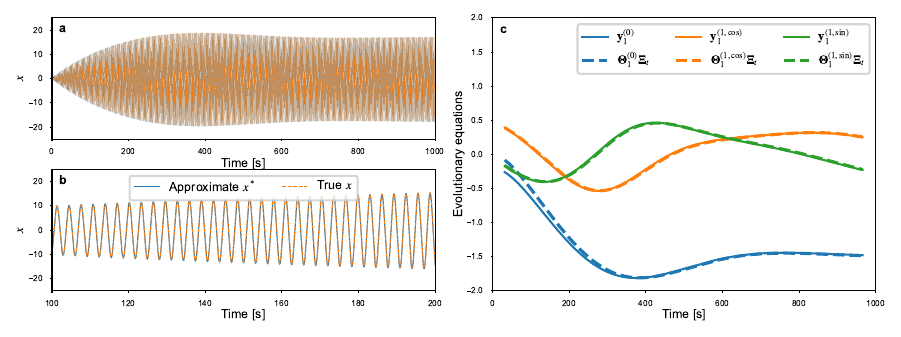}
    \caption{Benchmark 1D forced oscillator. Numerical validation of the Generalized Harmonic Balance (GHB) method: (a) Comparison between the ground-truth displacement $x$ (numerical integration) and the reconstructed $x^*$ obtained via harmonic superposition, see Eq.~\eqref{eq:WNO_solution0}; (b) zoomed-in view of the transient response highlighting the reconstruction accuracy; (c) validation of the evolutionary terms, comparing the analytical transformations with their numerical counterparts for various harmonic components.}
    \label{fig:theroyProvement}
\end{figure}

We evaluate the theoretical foundation of the proposed method, specifically focusing on the validity of slow-varying evolutionary variables and the generalized harmonic balance formulation. The fundamental assumption is that the response $x(t)$ of a weakly nonlinear forced oscillator can be accurately represented as a superposition of multiple harmonic components. Figure~\ref{fig:theroyProvement}(a) presents a comparison between the ground-truth response $x(t)$, obtained via numerical integration, and the approximated response $x^{*}(t)$ reconstructed using Eq.~\eqref{eq:WNO_solution0}. Based on the spectral characteristics identified in Figure~\ref{fig:numericaldata}, only the zero-order and first-order harmonic components are selected for the oscillator in Eq.~\eqref{eq:WNO_case1}. We neglect the second-order harmonic in this specific case. It is important to note that including insignificant harmonic components can degrade the inference results. The slowly varying evolutionary variables $a^{(0)}(t)$, $b^{(1)}(t)$, and $c^{(1)}(t)$ are extracted following the procedure described in Sec.~\ref{sec:method-filter}. As illustrated in the zoomed-in view in Figure~\ref{fig:theroyProvement}(b), the reconstructed signal $x^{*}(t)$ exhibits excellent agreement with the reference response $x(t)$, thereby confirming the robustness of the proposed approximation. 

Furthermore, the validation of the GHB-based reformulation (Eq.~\eqref{eq:GHB method_rewritten}) is presented in Figure~\ref{fig:theroyProvement}(c), comparing the terms derived from the evolutionary variables, with those computed by numerically integrating the weak nonlinearities $f(x,\dot{x})$ over a single period. The close match verifies the accuracy of the GHB method and establishes the equivalence between the original physical formulation (Eq.~\eqref{eq:WNO_start}) and the transformed evolutionary representation(Eq.~\eqref{eq:GHB method}). 

\subsection{Inferred equations}

\begin{figure}
    \centering
    \includegraphics[width=\linewidth]{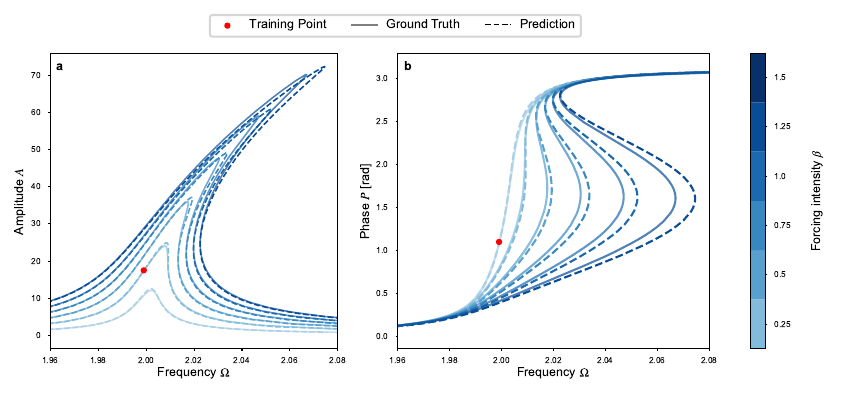}
    \caption{
Benchmark 1D forced oscillator. Comparison between predicted and reference frequency response functions under different excitation amplitudes. The proposed method accurately captures the hardening dynamical performance of oscillator.
}
\label{fig:caes1FRF}
\end{figure}

The proposed method is applied to the benchmark 1D oscillator to discover the governing equations 
and reconstruct its complete dynamic portrait, including frequency response curves (FRCs). 
To capture potential nonlinear behaviors, a candidate library $\mathbf{\Theta}$ is constructed, consisting of ten pure polynomial terms up to the fifth order and trigonometric functions ($\cos(\Omega t)$ and $\sin(\Omega t)$). Notably, cross-coupling terms between displacement and velocity are omitted to maintain model parsimony. The algorithm is implemented with the following hyper-parameters: the effective harmonic orders are selected as $[0, 1]$, the residual tolerance is set to $1\times 10^{-1}$, and the lowest contribution ratio is $5\times 10^{-2}$. A significant advantage of the proposed approach is its data efficiency: the exact equation structure is successfully identified using data from only a single excitation case (indicated by the red point in Figure~\ref{fig:caes1FRF}). The estimated coefficients, reported in Table~\ref{tab:case1results}, exhibit high precision relative to the reference values.

The proposed method successfully identifies the correct equation structure from the library by only one excitation case (red Point in Figure~\ref{fig:caes1FRF}). Estimated coefficients are reported in Table~\ref{tab:case1results}. These results demonstrate 
that the proposed approach achieves accurate and parsimonious discovery of the weakly 
nonlinear dynamics.

\begin{table}[]
\centering
\caption{Benchmark 1D forced oscillator. The "Reference" columns report the parameters used for the training ; in "MEv-SINDy" columns, we list the parameter values identified by our method.}
\begin{tabular}{cccc}
\hline
Coefficients & Relevant term      & Reference             & MEv-SINDy           \\ \hline
$\omega^2$   & $x$                & $4.0000$              & $4.0012$              \\
$c$          & $\dot{x}$          & $1.0000\times 10^{-2}$ & $1.0495\times 10^{-2}$ \\
$\alpha_1$   & $\dot{x}^2$        & $1.0000\times 10^{-2}$ & $1.0071\times 10^{-2}$ \\
$\alpha_2$   & $x^3$              & $1.0000\times 10^{-4}$ & $9.6691\times 10^{-5}$ \\
$\beta$            & $\cos(\Omega t)$ & $5.0000\times 10^{-1}$ & $5.0748\times 10^{-1}$ \\ \hline
\end{tabular}

\label{tab:case1results}
\end{table}

After the governing equation is identified, the system frequency response function (FRF) 
can be computed either by time integration or by a parameter continuation method. The FRF here 
describes the stationary oscillation amplitude of the state $x$ as a function of the excitation frequency 
$\Omega$. As shown in Figure~\ref{fig:caes1FRF}, the predicted FRF closely matches the 
ground-truth results for various excitation frequencies and amplitudes. It is worth emphasizing, although the governing equation is inferred using only a single excitation condition, the entire 
frequency-response space under arbitrary forcing conditions can subsequently be predicted with 
high accuracy.

\section{Applications}
\label{sec:highD_app}

To highlight the capabilities of the proposed approach, we investigate high-dimensional structural dynamical systems exhibiting large-amplitude vibrations and geometric nonlinearities. Micro-Electro-Mechanical Systems (MEMS) are typically actuated near their resonance frequencies and consequently undergo substantial deformations. These effects are further intensified by the fact that MEMS are monolithic devices often encapsulated in near-vacuum environments, where energy dissipation is minimal. As a result, the system dynamics are strongly nonlinear and characterized by complex coupled behaviors, showing highly nonlinear dynamical features that are rarely observed at the macro scale. Therefore, MEMS provide a representative and valuable platform for applying the proposed approach to evaluate its capability in predicting dynamical performance through the inferred governing equations.

In this section, we demonstrate the application of the proposed method to high-dimensional structural dynamical systems using synthetic data. The discussion begins with the formulation of the full-order model (FOM) based on the finite element discretization of the governing equations. Subsequently, the system is reduced through the Proper Orthogonal Decomposition (POD) technique to obtain a reduced-order representation that retains the dominant dynamical features. To illustrate the procedure and evaluate the performance of the method, two case studies are presented: a clamped–clamped beam, and a more complex MEMS micromirror device that demonstrates the scalability of the proposed approach.

\subsection{Formulation of full-order model}

The framework of structures subjected to large displacements and small strains is considered for full-order model simulations~\citep{holzapfel2002nonlinear}. This is the operating range of most microsystems since they are often actuated at resonance and large aspect ratios allow reaching displacements within the linear elastic range of the material. In this framework, the Saint Venant-Kirchhoff constitutive model is the most appropriate choice, and is given by:

\begin{equation}\label{eq:Saint Venant-Kirchhoff}
    \mathbf{S}=\mathcal{A}:\mathbf{E}, \qquad \mathbf{E}=\frac{1}{2} (\nabla \mathbf{d}+ \nabla^T\mathbf{d}+ \nabla^T\mathbf{d}\cdot\nabla \mathbf{d})
\end{equation}
where $\mathbf{S}$ is the second Piola-Kirchhoff strain tensor, $\mathcal{A}$ is the fourth-order elasticity tensor and $\mathbf{E}$ is the Green-Lagrangian Strain tensor. Here we denote by $\mathbf{d}$ the displacement field and by $\nabla(\cdot)$ the (material) gradient defined with respect to the reference configuration. The weak form of the linear momentum conservation law is:

\begin{equation}\label{eq:linear momentum}
    \int_{\mathit{\Omega}_0} \rho_0 \mathbf{\ddot{d}}\cdot\mathbf{w}d\mathit{\Omega}_0+\int_{\mathit{\Omega}_0}\mathbf{P}[\mathbf{d}]:\nabla^T\mathbf{w}d\mathit{\Omega}_0=\int_{\mathit{\Omega}_0}\rho_0\mathbf{F}\cdot\mathbf{w}d\mathit{\Omega}_0+\int_{S_{T0}}\mathbf{f}\cdot\mathbf{w}dS_0,\qquad \forall\mathbf{w}\in H_0^1(\mathit{\Omega}_0)
\end{equation}
where the integrals are expressed in the reference configuration $\mathit{\Omega}_0$. Here, $\rho_0$ denotes the material density, $\mathbf{P}[\mathbf{d}]=(1+\nabla \mathbf{d})\cdot \mathbf{S}$ is the first Piola-Kirchhoff stress tensor, $\mathbf{F}$ is the body force per unit mass, $\mathbf{f}$ is the surface tractions prescribed on the surface $S_{T0}$ and $\mathbf{w}$ is the test velocity selected in $H_0^1(\mathit{\Omega_0})$, i.e. the space of functions with finite energy that vanish on the portion $S_U\subset \partial\mathit{\Omega_0}$ where Dirichlet boundary conditions are prescribed. 

By applying spatial discretization to Eq.~\eqref{eq:linear momentum}, for instance through the finite element method, and incorporating a Rayleigh damping model, the system can be transformed into a set of coupled nonlinear ordinary differential equations of the following form:
\begin{equation}\label{eq:spatialDiscretization}\mathbf{M\ddot{D}+C\dot{D}+KD+G(D,D)+H(D,D,D)}=\mathbf{B}(\mathbf{D,\beta},\Omega,t),\qquad t\in(0,T)
\end{equation}
where the vector $\mathbf{D}\in \mathbb{R}^k$ collects all the unknown displacement nodal values, $\mathbf{M}\in\mathbb{R}^{n\times n}$ is the mass matrix, $\mathbf{C}=(\omega_0/Q)\mathbf{M}$ is the Rayleigh model mass proportional damping matrix, defined with respect to a reference eigenfrequency $\omega_0$ and a quality factor $Q$. $\mathbf{B}\in \mathbb{R}^n$ is the nodal force vector which depends on the actuation intensity parameters $\beta$, the angular frequency of actuation $\Omega$ and in general also on $\mathbf{D}$. The internal force vector has been exactly decomposed in linear, quadratic, and cubic power terms of the displacement: $\mathbf{K}\in\mathbb{R}^{n\times n}$ is the linear stiffness matrix, while $\mathbf{G}\in\mathbb{R}^n$ and $\mathbf{H}\in\mathbb{R}^n$ are vectors given by monomials of second and third order, respectively. The components of these vectors can be expressed using an indicial notation: $G_i=\sum_{j,k=1}^nG_{ijk}D_jD_k,\: H_i=\sum_{j,k,l=1}^nH_{ijkl}D_jD_kD_l,\:i=1,\cdots,n.$

Eq.~\eqref{eq:spatialDiscretization} represents the high-fidelity full-order model (FOM) which depends on the input parameters $\beta$ and $\Omega$. It should be recalled that in resonating MEMS FRCs are an important design tool in which a selected quantity, like the midspan deflection of a beam or the rotation of a micromirror, is plotted versus $\Omega$ for different $\beta$. 
Indeed, resonators operate close to a reference frequency where the behavior should be predictable. However, approximating and reconstructing different FRCs for high dimensional dynamical system is computationally demanding, making the use of full-order models for this task largely impractical

\subsection{Proper Orthogonal Decomposition}

To substantially accelerate the computationally expensive simulations and obtain a more interpretable and parsimonious set of governing equations, we propose a reduction technique that simultaneously decreases the problem dimensionality and introduces a new set of coordinates better adapted to capture the system dynamics within only a few dominant modes. The proposed reduction technique is based on Proper Orthogonal Decomposition (POD), a model reduction method that constructs an ordered set of basis functions according to their energy content. By projecting the full-order data onto a limited number of dominant high-energy modes, a reduced-order representation of the system can be efficiently obtained~\citep{rowley2004model}. The snapshots of the FOM solutions are collected to generate a matrix $\mathbf{X}\in\mathbb{R}^{p\times k}$, whose $p$ denotes the time stamps of solutions. 

Next, the Singular Value Decomposition (SVD) of the matrix $\mathbf{X}$ is computed: $\mathbf{X}=\mathbf{U\Sigma V^T}$, where the columns of the orthonormal matrix $\mathbf{U}\in \mathbb{R}^{k\times k}$ are left singular vectors, often called Proper Orthogonal Modes (POMs)~\citep{abbott2005model,lu2019review} or spatial modes, the columns of $\mathbf{V}\in \mathbb{R}^{p\times p}$ give the temporally evolving coefficients, and $\mathbf{\Sigma}\in\mathbb{R}^{k\times p}$ is a diagonal matrix representing the singular values of the matrix $\mathbf{X}$ ordered from the largest to the smallest conventionally. 
In particular, the rank of $\mathbf{X}$ is equal to the number of nonzero singular values, and the optimal rank-$\hat{k}$ approximation $\hat{\mathbf{X}}$ of $\mathbf{X}$ is given by the rank-$\hat{k}$ SVD truncation $\hat{\mathbf{X}}=\sum_{i=1}^{\hat{k}}\sigma_i\mathbf{U}_i\mathbf{V}_i^T$, in which $\sigma_i$ is the $i$th singular value contained in the diagonal of $\mathbf{\Sigma}$, and $\mathbf{U}_i,\mathbf{V}_i$ are the $i$th column of $\mathbf{U}$ and $\mathbf{V}$, respectively. 
In this work, $\hat{k}=3$, which gives a reduced-order model of dimension 3. The reduced order dynamical coordinates $\mathbf{\Lambda}_i=\sigma_i\mathbf{V}_i^T,\quad i=1,2,3$.




\subsection{High-dimensional case 1: Beam MEMS resonator}

The doubly clamped beam is investigated as a representative example of a high-dimensional nonlinear structural system, where a straight beam-type MEMS resonator is excited near its resonance frequency. This configuration captures the essential dynamics of micro-resonators~\citep{zega2020numerical} in which geometric nonlinearities play a significant role in shaping the system’s response.

\begin{figure}[h]
    \centering
    \includegraphics[width=1\linewidth]{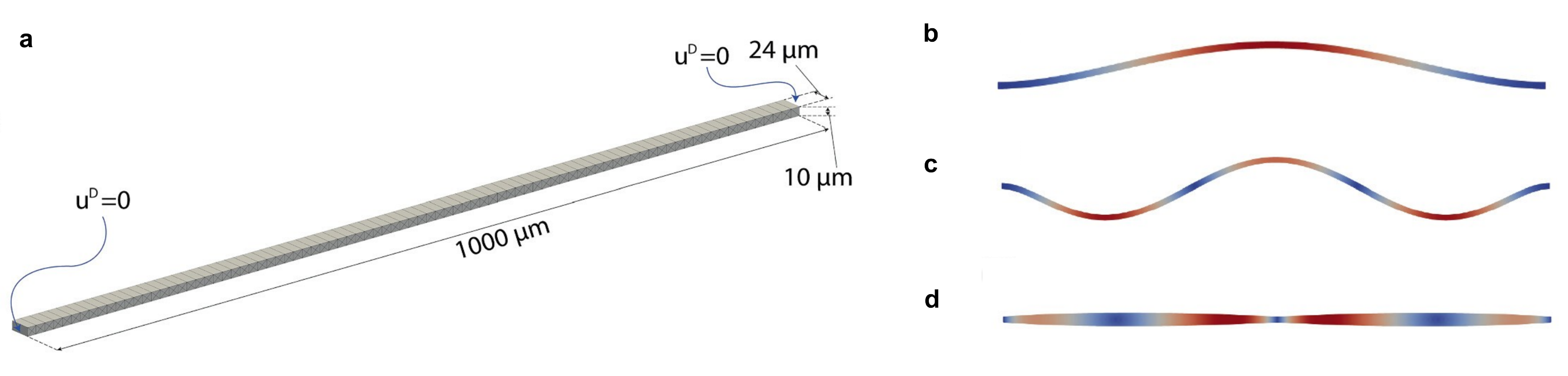}
    \caption{Schematic representation of the beam MEMS resonator.
    (a) Geometry and mesh of doubly clamped beam MEMS resonator,
    (b-d) First three most energetic POD modes obtained from SVD
    }
    \label{fig:Microbeam_setup}
\end{figure}

The doubly clamped beam in this case has length $L=1000\,\mu$m with rectangular cross-section of dimensions $10\mu \text{m}\times 24\mu \text{m}$, as shown in Figure~\ref{fig:Microbeam_setup}a. The beam is fabricated from isotropic polysilicon~\citep{corigliano2004mechanical}, with density $\rho=2330$\,kg/m$^3$, Young modulus $E=167$\,GPa and Poisson coefficient $\nu=0.22$. Dirichlet boundary conditions are enforced on the two opposite sides of the beam. The quality factor is assumed equal to $Q=50$. The device is excited by a body load $\mathbf{F}=\mathbf{M}\bm{\phi}_1\beta\cos(\Omega t)$ proportional to the first eigenmode $\bm{\phi}_1$, with $\mathbf{M}$ mass matrix and $\beta$ load multiplier. The device vibrates according to its first bending mode at $\omega_0 = 0.5475 \text{rad}/\mu \text{s}$. In order to keep the FOM computational time at a reasonable level, a rather coarse mesh with 2607 nodes has been employed. To effectively infer the underlying dynamics of beam MEMS resonator, the POD modes are built retaining in the linear trial space the first three most energetic POD modes, depicted in Figure~\ref{fig:Microbeam_setup}b-d, respectively. This number of POD modes represents the minimum required to achieve a good accuracy while maintaining a computationally efficient model~\citep{conti2023reduced}.

\begin{figure}[h]
    \centering
    \includegraphics[width=1\linewidth]{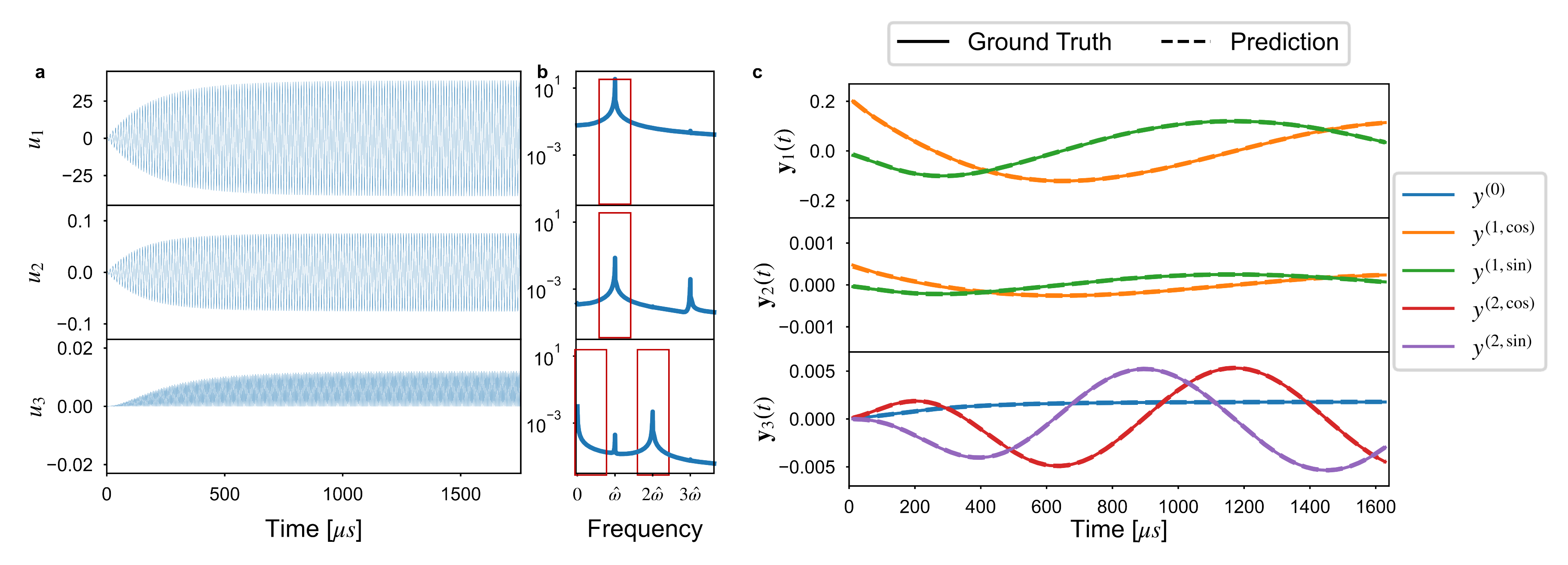}
    \caption{
    Beam MEMS resonator training data in time, frequency and evolutionary domain. The model is trained using a single parameter set ($\beta = 0.25, \Omega = 0.551$). 
    (a) Time series of the POD modal coordinates;
    (b) Frequency spectra, where red boxes indicate the identified effective harmonic orders; 
    (c) Comparison of slowly varying variables for the first three POD modes, showing excellent agreement between the ground truth (solid) and prediction by Eq.\eqref{eq:Beam}(dashed).
    }
    \label{fig:beam fitting}
\end{figure}

To infer the underlying dynamics of the MEMS resonator, training data are collected from a single simulation case with a load multiplier $\beta = 0.25$ and excitation frequency $\Omega = 0.551$. Figure~\ref{fig:beam fitting} illustrates the characteristics of the training set in the frequency, time and evolutionary domains. Specifically, Figure~\ref{fig:beam fitting}a displays the time-history responses of the first three POD modes, while Figure~\ref{fig:beam fitting}b shows their corresponding frequency spectra. The red boxes in the spectra highlight the effective harmonic orders, confirming that the system response is predominantly composed of multiple discrete frequencies. Using the proposed filtering procedure (Sec 2.4), the slow varying evolutionary variables for the POD modes are extracted, as shown in Figure~\ref{fig:beam fitting}c. Building upon these extracted variables, our approach successfully identifies a parsimonious set of governing equations that govern the reduced-order space from a candidate library $\mathbf{\Theta}$ consisting of ten pure polynomial terms up to the fifth order and trigonometric functions ($\cos(\Omega t)$ and $\sin(\Omega t)$), as explicitly formulated in Eq.~\eqref{eq:Beam}. The identified model is capable of accurately reconstructing the slow-varying evolutionary variables (Figure~\ref{fig:beam fitting}c dashed line), confirming that the identified nonlinear terms correctly capture the weakly nonlinear effects across different harmonic orders. Specifically, these identified equations capture the linear restorative forces alongside the geometric nonlinearities inherent in the large-deflection regime of the MEMS beam.

\begin{equation}
\begin{aligned}
\left\{
\begin{array}{lll}
    \ddot{u}_1\!+\!2.998\!\times\!10^{-1} u_1\!+\!2.730\times 10^{-6}u_1^3+1.100\times 10^{-2} \dot{u}_1&=1.902\beta\cos(\Omega t)\\
    \ddot{u}_2\!+\!3.006\!\times\!10^{-1} u_2 \!-\! 1.468\!\times\! 10^{-6}u_1\!+\!5.453\!\times\! 10^{-9}u_1^3\!+\!2.450\! \times\! 10^{-5}\dot{u}_1\!\!&=4.202\!\times\!10^{-3}\beta\cos(\Omega t)\\
    \ddot{u}_3\!+\!3.006\!\times\!10^{-1} u_3\!+\!2.402\!\times \!10^{-6}u_1^2-1.574\!\times\! 10^{-5} \dot{u}_1^2&=0\\
        \end{array}
\right.
\end{aligned}
\label{eq:Beam}
\end{equation}

\begin{figure}[h]
    \centering
    \includegraphics[width=1\linewidth]{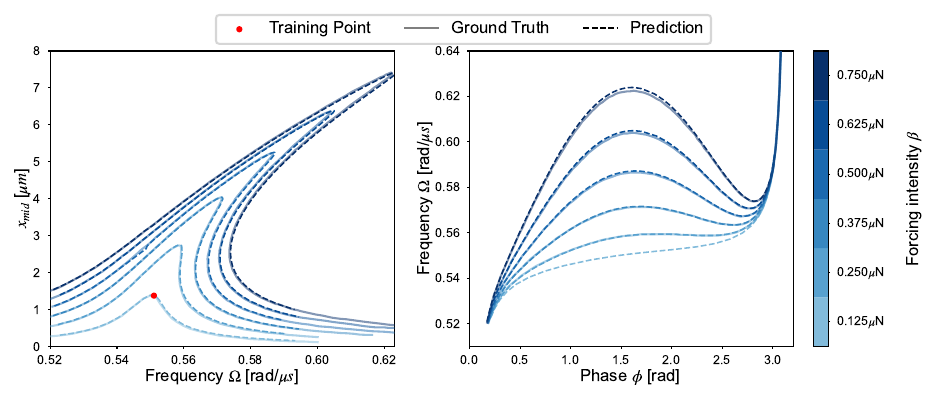}
    \caption{FRCs of beam MEMS resonator. Comparison between predicted and reference frequency response functions under different excitation amplitudes (left: amplitude; right: phase). The proposed method accurately captures the hardening dynamical performance of beam MEMS resonator.}
    \label{fig:Beam_FRF}
\end{figure}

The resulting model provides a compact yet precise representation of the resonator complex dynamics, enabling efficient prediction of its complete dynamic portrait while significantly reducing the computational burden compared to the full-order finite element model. Specifically, to evaluate the generalization capability of the identified equations, we predict the global FRCs of the beam MEMS resonator under various excitation intensities. As illustrated in Figure~\ref{fig:Beam_FRF}, the predicted FRCs (dashed line) demonstrate remarkable agreement with the reference FOM results (solid line) across entire investigated range of forcing amplitudes ($\beta\in[0.125,0.750]\mu$N), with the left and right panels presenting the amplitude-frequency and frequency-phase curves, respectively. In particular, 
the identified model successfully captures the nonlinear hardening behavior, characterized by the rightward tilting of the resonance peaks as the excitation amplitude increases. Notably, while the governing equations were inferred using data from only a single training point (the red dot at $\beta=0.25,\Omega=0.551$), the model accurately extrapolates the complex bifurcation characteristics and phase transitions for forcing levels significantly higher than the training condition. This high-fidelity match underscores the physical consistency and robustness of the proposed identification framework for high-dimensional nonlinear structural systems.



\subsection{High-dimensional case 2: MEMS micromirror}

Scanning micromirrors have experienced rapid growth in recent years, driven by their successful implementation in a wide range of applications, from pico-projectors for Augmented Reality (AR) displays to three-dimensional (3D) scanning systems for Light Detection and Ranging (LiDAR) technologies. Because of the inertial and geometrical effects triggered by large rotations, micromirrors are intrinsically nonlinear and the prediction of their dynamic behavior is essential to guarantee a proper design and control of the mirror during the online stage. Specifically, maintaining motion stability is critical for line scanning performance, making the correct characterization of hardening and softening effects a priority.

\begin{figure}[h]
    \centering
    \includegraphics[width=0.85\linewidth]{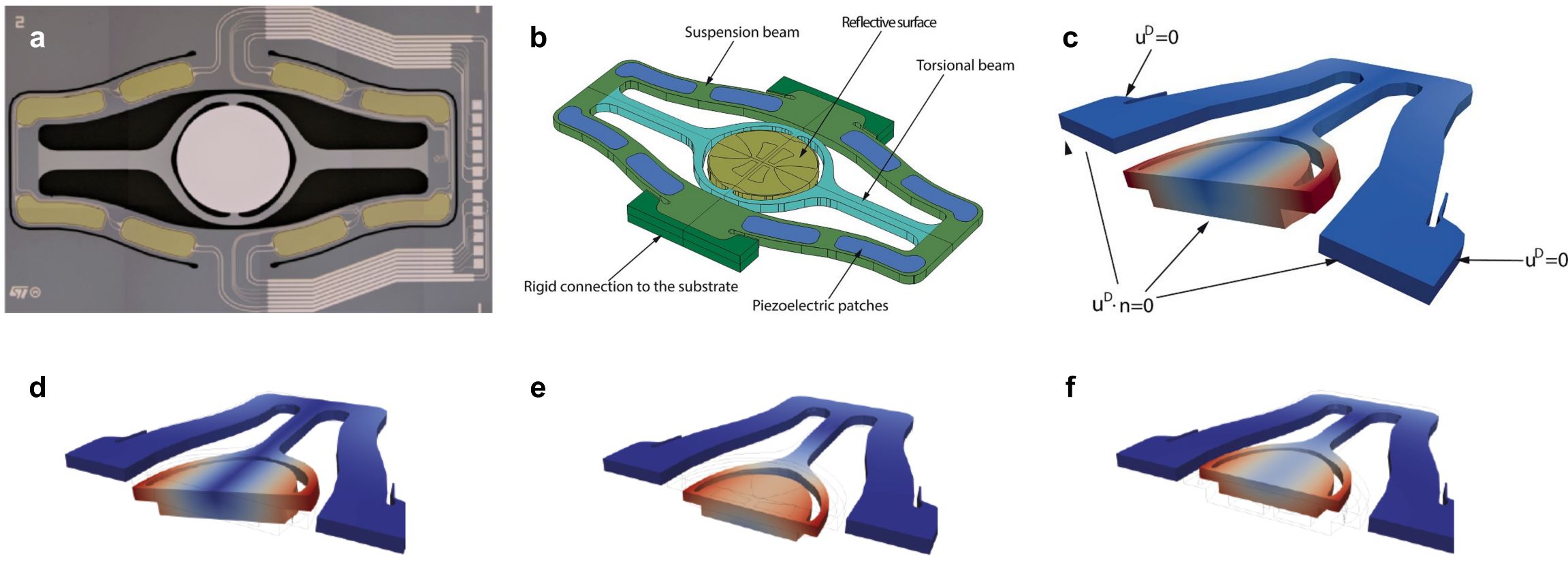}
    \caption{
    Schematic representation of the MEMS micromirror: 
    (a) Optical picture of the Micromirror;
    (b) Schematic view of the layout with few details on the device components;
    (c) Third (torsional) eigenmode that is actuated during operations and the boundary conditions used in the simulation of half of the device;
    (d-f) First three most energetic POD modes obtained from SVD.
    }
    \label{fig:setupmirror}
\end{figure}

The mirror considered, fabricated by STMicroelectronics, is illustrated in Figure~\ref{fig:setupmirror}. The mirror plate is suspended to a gimbal connected with a torsional beam along the rotation axis and two suspension beams on each side. The mirror is assumed to be made of isotropic polysilicon~\citep{corigliano2004mechanical}, with density $\rho=2330kg/m^3$, Young modulus $E=167 GPa$ and Poisson coefficient $\nu=0.22$. Exploiting structural symmetry,, only half of the micromirror is modeled with the FEM and a total of 9723 dofs. The Dirichelet boundary conditions are imposed in the substrate (the dark green areas in Figure~\ref{fig:setupmirror}b) and on the symmetry plane to enforce the symmetry conditions. Given the high stiffness of the central mirror plate, its rotation angle is adopted as the primary state variable for both the time-domain response and the frequency response function (FRF) analysis.

\begin{figure}[h]
    \centering
    \includegraphics[width=1\linewidth]{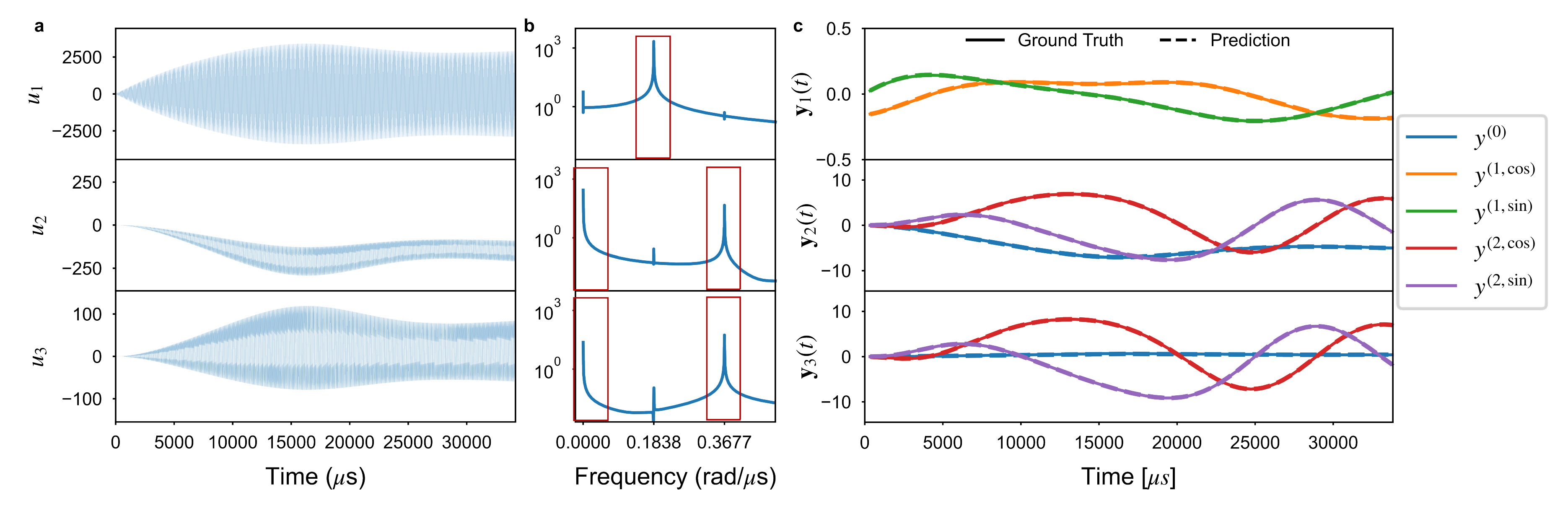}
    \caption{
    MEMS micromirror training data in time, frequency and evolutionary domains: The model is trained using a single parameter set ($\beta = 3, \Omega = 0.183841$). 
    (a) Time series of the POD modal coordinates;
    (b) Frequency spectra, where red boxes indicate the identified effective harmonic orders; 
    (c) Comparison of slowly varying variables for the first three POD modes, showing excellent agreement between the ground truth (solid) and prediction by Eq.~\eqref{eq:Mirror}(dashed).    
    }
    \label{fig:mirror_fitting}
\end{figure}

To characterize the complex nonlinearities of the scanning micromirror, the training set of the MEv-SINDy framework is derived from a single simulation trajectory with a load multiplier $\beta = 3$ and an excitation frequency $\Omega = 0.183841$. Figure~\ref{fig:mirror_fitting} illustrates the sampled data across the time, frequency and evolutionary domains. Unlike the beam case, the torsional tilting of the mirror triggers a richer multi-harmonic spectrum, as evidenced by the discrete frequency peaks in Figure~\ref{fig:mirror_fitting}b. The filtering procedure effectively isolates the evolutionary variables for the underlying manifold (Figure~\ref{fig:mirror_fitting}c), providing a clean foundation for sparse regression. The algorithm yields a compact set of reduced-order equations (Eq.~\eqref{eq:Mirror}) that faithfully reproduce the training signals (see the comparison between continuous and dashed lines in Figure~\ref{fig:mirror_fitting}c). Notably, the identified model successfully decouples the inertial nonlinearities and cross-modal interactions inherent in the gimbal-suspension assembly.

\begin{equation}
\begin{aligned}
\left\{
\begin{array}{lll}
    \ddot{u}_1+3.383\times10^{-2} u_1-1.180\times 10^{-11}u_1^3+1.865\times 10^{-4} \dot{u}_1&=1.076\times 10^{-1}\beta\cos(\Omega t)\\
    \ddot{u}_2+3.374\times10^{-2} u_2 - 1.103\times 10^{-7}u_1^2+3.912\times 10^{-5}\dot{u}_1^2&=0\\
    \ddot{u}_3+3.374\times10^{-2} u_3- 9.127\times 10^{-7}u_1^2+2.392\times 10^{-5} \dot{u}_1^2&=0\\
    \end{array}
\right.
\end{aligned}
\label{eq:Mirror}
\end{equation}

\begin{figure}[h]
    \centering
    \includegraphics[width=1\linewidth]{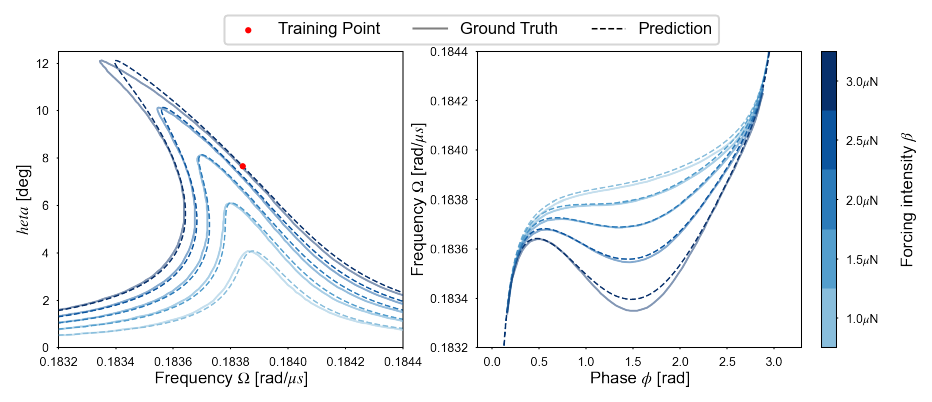}
    \caption{FRCs of MEMS micromirror. Comparison between predicted and reference frequency response functions under different excitation amplitudes (left: amplitude; right: phase). The proposed method accurately captures the softening dynamical performance of MEMS micromirror.}
    \label{fig:mirror_FRF}
\end{figure}

The robustness of the identified model is further validated by predicting the global Frequency Response Functions (FRCs). As shown in Figure~\ref{fig:mirror_FRF}, the MEv-SINDy framework accurately extrapolates the system behavior across a wide range of forcing intensities. Of particular interest is the model ability to capture the softening behaviors, which is a critical factor for ensuring scanning stability in LiDAR and AR applications. The seamless agreement between the ROM predictions and the FOM ground truth—despite the model being trained on a single localized point—underscores the physical consistency of the discovered dynamics.

\section{Robustness experiments}

While the preceding cases demonstrate the high fidelity of the MEv-SINDy framework in identifying complex structural dynamics, a systematic assessment of its robustness is essential to define its practical operational boundaries. This section investigates the sensitivity and reliability of the proposed method under varying data conditions, focusing on two critical aspects. First, we examine the influence of training point selection on global predictive accuracy, specifically investigating whether the location of the training sample affects the model ability to extrapolate the system global response. Second, we evaluate the scaling of prediction precision with the number of training points. By comparing the performance of models trained on sparse versus enriched datasets, we highlight our approach primary advantage: the ability to achieve high-dimensional ROM discovery with minimal data requirements, thereby significantly reducing the computational overhead inherent in traditional data-driven modeling.

\subsection{Influence of training point selection} 

To provide a comprehensive assessment of MEv-SINDy's reliability, we conduct extensive testing across two benchmark datasets: the beam MEMS resonator (56 distinct operating points, featuring specific forcing frequency and amplitude) and the MEMS micromirror (22 distinct operating points). For each individual point in these datasets, the governing equations are first identified via the proposed framework. Subsequently, the full-range frequency responses are reconstructed using MATCONT (as detailed in Sec.~2.7), a sophisticated parameter continuation solver. To quantify the fidelity of the identified models, we utilize the Mean Chamfer Distance of Frequency Response Curves (MCDRC) as our primary metric (as detailed in Sec.~2.8). Specifically, a score below MCDRC < 0.1 indicates that the identified model has successfully captured the essential nonlinear characteristics across the entire dynamical regime.

\begin{figure}[h]
    \centering
    \includegraphics[width=1\linewidth]{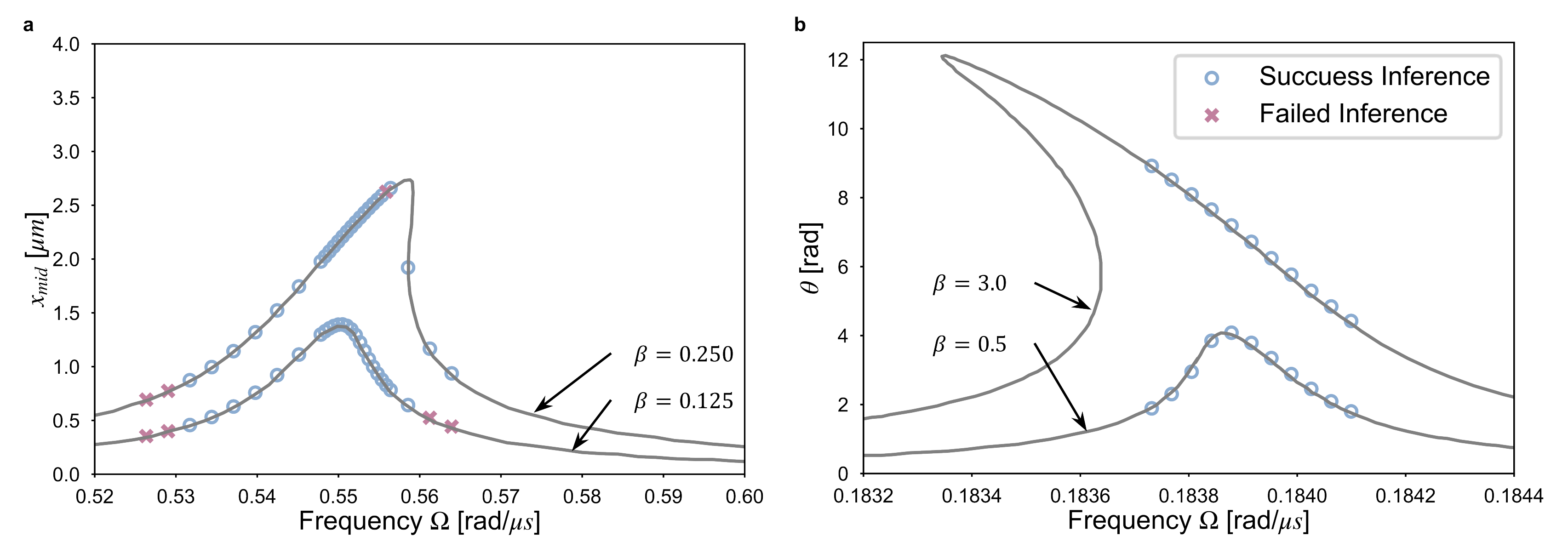}
    \caption{
    Robustness evaluation of proposed approach across various training conditions. The predictive performance is assessed by identifying the model using a single training point and evaluating its global extrapolation capability. Blue circles indicate a successful inference (MCDRF < 0.1), while red crosses denote a failed inference.
    (a) Beam MEMS resonator: results for 56 different training points, covering two excitation amplitudes $\beta\in\{0.125,0.250\}$ and 28 frequencies $\Omega$ ranging from $0.526$ to $0.564$ rad/$\mu$s. A total of 51 training points yield successful global predictions.
    (b) MEMS micromirror: results for 22 distinct training points, covering two excitation amplitudes $\beta \in \{0.5, 3.0\}$ and 11 frequencies $\Omega$ ranging from $0.1837$ to $0.1841$ rad/$\mu$s. All 22 training points achieve successful inference.
    }
    \label{fig:robust-trainingpoints}
\end{figure}

 In particular, the reliability of the MEv-SINDy framework is systematically evaluated by analyzing the influence of training point selection on global predictive accuracy. As illustrated in Figure 13, the framework exhibits remarkable robustness across diverse operating regimes. For the beam MEMS resonator, 51 out of 56 candidate training points yield successful global inferences (Figure~\ref{fig:robust-trainingpoints}a), while for the MEMS micromirror, the framework achieves a 100\% success rate across all 22 tested points (Figure~\ref{fig:robust-trainingpoints}b). Notably, accurate global ROMs are identified when the training data is sampled from the strongly nonlinear regimes characterized by significant resonance tilting.

\begin{figure}[h]
    \centering
    \includegraphics[width=1\linewidth]{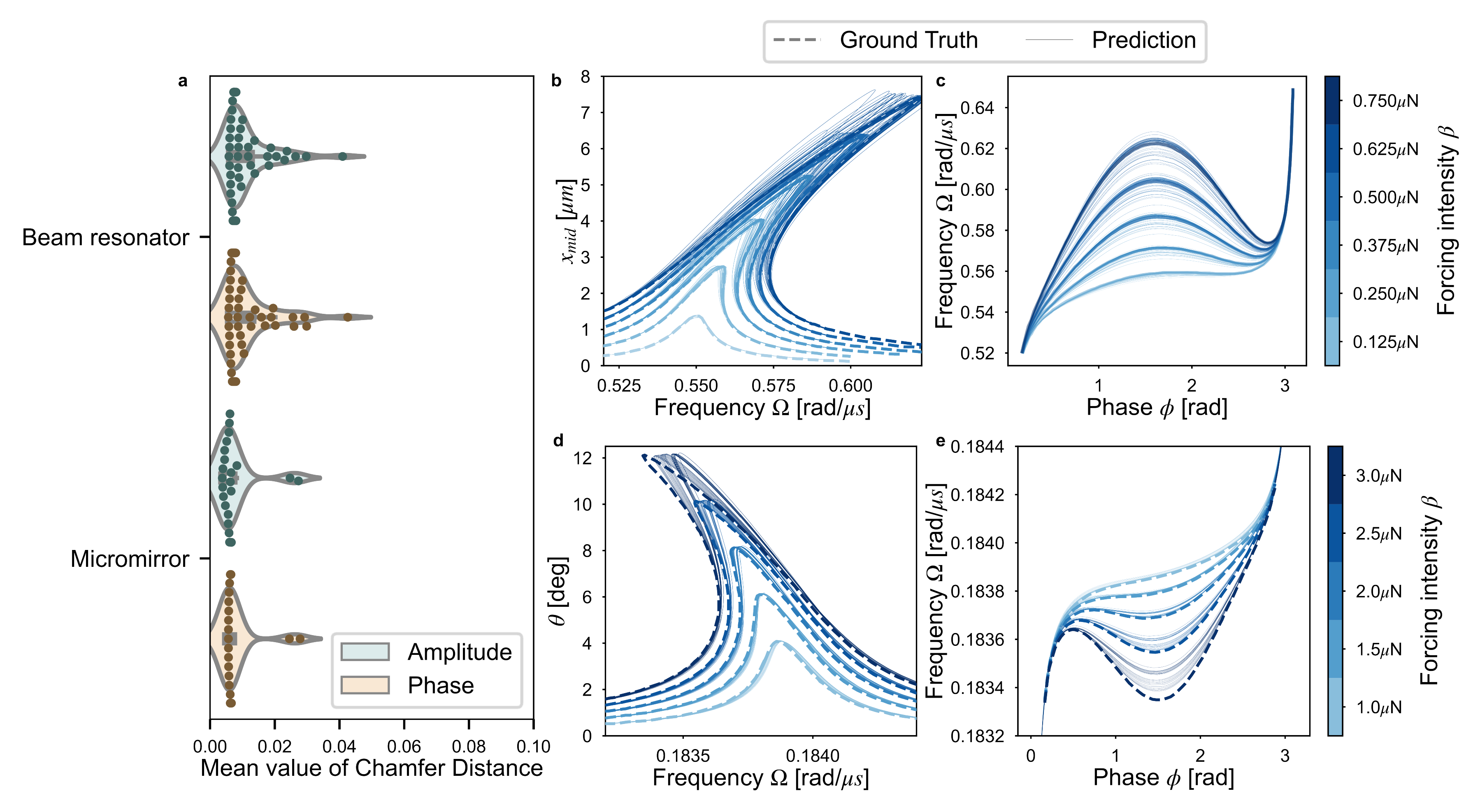}
    \caption{
    Quantitative and qualitative error analysis of FRF predictions.
    (a) MCDRF distribution across all successful inferences; violin plots show high error concentration near zero for both devices. 
    (b–c) Qualitative comparison between Ground Truth (dashed) and predictions (solid) for the beam resonator, and 
    (d–e) for the micromirror. The consistent overlap across various training conditions highlights the framework’s minimal sensitivity to training data selection.
    }
    \label{fig:robust-metrics}
\end{figure}

This consistency is further quantified in Figure~\ref{fig:robust-metrics}a, where the MCDRC distribution for all successful cases is shown to be highly concentrated near zero. The narrow spread of these violin plots underscores that the identification fidelity is largely invariant to the specific choice of training conditions. Qualitatively, Figures~\ref{fig:robust-metrics}b–e confirm this high fidelity; the FRC curves reconstructed from various localized training points demonstrate a nearly perfect overlap with the ground-truth simulations. The framework faithfully reproduces complex nonlinear phenomena, including the softening behavior of the beam and the dual hardening-softening transitions in the micromirror. Collectively, these results demonstrate that MEv-SINDy effectively extracts the underlying physical operators rather than merely overfitting localized data, enabling reliable global extrapolation from minimal, localized measurements.

To further enhance the reliability of the MEv-SINDy framework within a one-shot learning context, it is imperative to investigate the specific conditions under which identification fails. The five failure cases observed in the beam MEMS resonator (Figure~\ref{fig:robust-metrics}a) can be categorized into two distinct modes:

\begin{enumerate}
    \item \textbf{Limited nonlinear effects}: Several failed points are located at frequencies far from the system natural frequency. In these near-linear regimes, the nonlinear contributions to the dynamical response are extremely small. Consequently, the sparse regression algorithm only discovers the linear damping for the first POD mode. While the resulting model can predict a precise dynamical response within the local linear regime, it lacks the cubic stiffness nonlinearity required for global extrapolation. This underscores that for robust one-shot learning, the training point must be selected from a region where nonlinear effects are sufficiently prominent to be accounted by the sparse identification process.
    \item \textbf{Incorrect identified dynamics}: Occasionally, identification failure occurs because the discovered model converges to an incorrect dynamical equation, leading to an erroneous global FRF prediction. Fortunately, such discrepancies are readily detectable during the validation phase; by comparing the model's predicted time-history against the original sampled trajectory, one can easily identify mismatches in time domain. To rectify this, two primary strategies can be employed: fine-tuning the hyperparameters (such as the sparsity threshold) to re-infer the true underlying dynamics or re-selecting a more informative training point that better characterizes the system's nonlinear landscape.
\end{enumerate}

\subsection{Multi-point training and frequency normalization}

While the MEv-SINDy framework demonstrates high reliability in one-shot learning, incorporating data from multiple excitation conditions can further stabilize the identification process. However, a fundamental mathematical discrepancy arises when attempting to directly merge datasets from different operating points. In the physical domain, the underlying system parameters (Eq.~\eqref{eq:WNO_start}) are intrinsic constants across different excitation cases. 
The linear frequency term requires special consideration. Its physical coefficient $\omega_0^2$ is split into two distinct parts. The first part is the observed frequency $\hat{\omega}$ obtained via FFT (Eq.~\eqref{eq:fft}). The second part is a small fine-tuning coefficient identified through the regression process (Eq.~\eqref{eq:multitargets}).
So the governing equations in our identification framework (Eq.~\eqref{eq:multitargets}) are constructed based on the observed response frequency $\hat{\omega}$, which is determined via FFT for each specific excitation case. Because $\hat{\omega}$ varies across different forcing conditions, the mathematical representation of the linear frequency in $\bm{\Theta}(t)$ becomes coupled with the local operating point. Consequently, Eq.~\eqref{eq:multitargets} yields inconsistent coefficients for the linear frequency term across datasets, preventing a straightforward least-squares concatenation.

To resolve this, we introduce a frequency-normalization step after reformulate the evolutionary library. By selecting a reference frequency $\bar{\omega}$ (e.g., the mean of all $\hat{\omega}$), we apply a compensation term to the target vector $\bm{y}(t)$. This ensures the linear frequency term is mapped to a consistent basis, effectively decoupling the physical parameters from the local observation frequency. The corrected target vector $\bm{y}_{corr}(t)$ replaces the original vector $\bm{y}$ in Eq.~\eqref{eq:multitargets}. This substitution ensures that the regression targets are consistent across all training sets.
\begin{equation}\label{eq:FrequencyCorr}
    \bm{y}_{corr}(t) = \bm{y}(t)-(\hat{\omega}^2-\bar{\omega}^2)\bm{\Theta}_{x}(t)
\end{equation}
where $\bm{\Theta}_{x}$ stands for the candidate function $x$ in the evolutionary library:
%
\begin{equation}
\bm{\Theta}_{x}(t)=\begin{bmatrix}
\bm{\Theta}_{x}^{(0)}(t)\\ 
\bm{\Theta}_{x}^{(1,\mathrm{cos})}(t)\\   
\bm{\Theta}_{x}^{(1,\mathrm{sin})}(t)\\ 
\bm{\Theta}_{x}^{(2,\mathrm{cos})}(t)\\
\bm{\Theta}_{x}^{(2,\mathrm{sin})}(t)\\
\vdots
\end{bmatrix}=\begin{bmatrix}
-\frac{\hat{\omega}_i}{2\pi}\int_{t-\pi/\hat{\omega}_i}^{t+\pi/\hat{\omega}_i}x(s)ds\\ 
-\frac{\hat{\omega}_i}{\pi}\int_{t-\pi/\hat{\omega}_i}^{t+\pi/\hat{\omega}_i}x(s)\cos(\hat{\omega}_is)ds\\   
-\frac{\hat{\omega}_i}{\pi}\int_{t-\pi/\hat{\omega}_i}^{t+\pi/\hat{\omega}_i}x(s)\sin(\hat{\omega}_is)ds\\ 
-\frac{\hat{\omega}_i}{\pi}\int_{t-\pi/\hat{\omega}_i}^{t+\pi/\hat{\omega}_i}x(s)\cos(2\hat{\omega}_is)ds\\
-\frac{\hat{\omega}_i}{\pi}\int_{t-\pi/\hat{\omega}_i}^{t+\pi/\hat{\omega}_i}x(s)\sin(2\hat{\omega}_is)ds\\
\vdots
\end{bmatrix}=\begin{bmatrix}
-a_0(t)\\ 
-b_1(t)\\   
-c_1(t)\\ 
-b_2(t)\\
-c_2(t)\\
\vdots
\end{bmatrix},
\label{eq:multitheta}
\end{equation}

\begin{figure}
    \centering
    \includegraphics[width=1\linewidth]{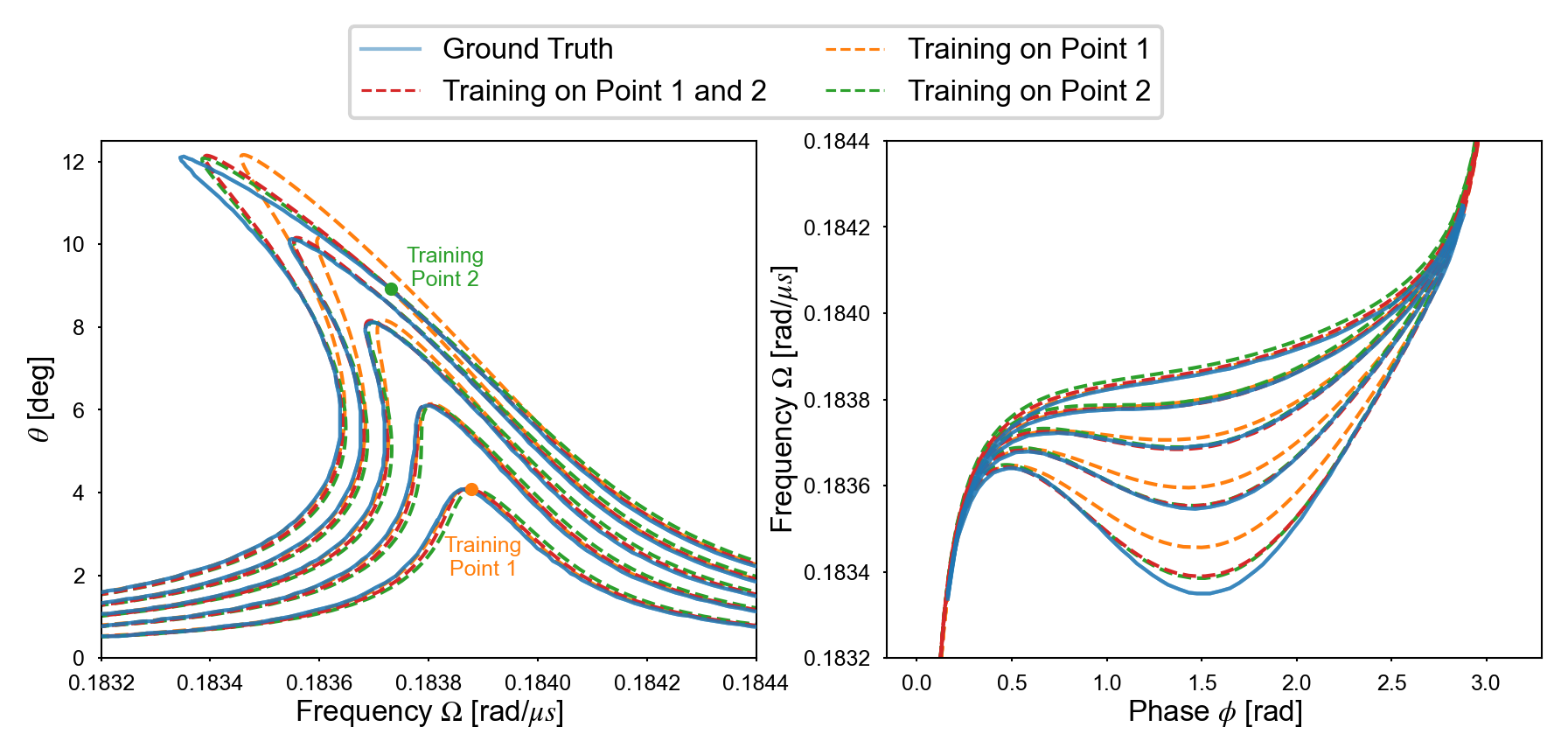}
    \caption{Comparison of Frequency Response Curves (FRCs) identified from different training sets. (left) Amplitude-frequency response; the forcing amplitudes increase from bottom to top. (right) Phase-frequency response; the forcing amplitudes decrease from bottom to top.}
    \label{fig:FRF_comparsion_MP}
\end{figure}

The efficacy of this multi-point approach is demonstrated in Figure~\ref{fig:FRF_comparsion_MP}. A comparison of the single-point training results reveals a trade-off in predictive performance: the model trained on Point 1 provides a better fit for low-amplitude excitation cases, whereas the model trained on Point 2 exhibits higher fidelity for high-amplitude regimes. This suggests that localized data, while accurate within its own regime, may not fully capture the system sensitivity across the entire forcing range. By implementing the frequency-normalization terms defined in Eq.~\eqref{eq:FrequencyCorr}, we can combine the complementary advantages of both points. The composite model (Training on Points 1 and 2) effectively fuses the ``low-amplitude'' and ``high-amplitude'' information into a single, unified ROM. As a result, the identified model achieves a superior and more consistent match with the ground-truth FRC under all tested conditions. This shows that our framework can successfully integrate disparate datasets to overcome the limitations of localized sampling, ensuring stable and high-fidelity global predictions.

\section{Conclusion}

Learning the complete dynamic portrait of weakly forced oscillators typically requires large training datasets. This is true especially when increasing popular deep learning are applied. In this study, we proposed a one-shot method to learn frequency-response curves from a single excitation configuration or, in other words, from a single traning point. To infer weakly nonlinear effects, we modified the Sparse Identification of Nonlinear Dynamic (SINDy) algorithm accounting for an evolution-based learning framework (MEv-SINDy). We shifted the focus from autonomous single-frequency dynamics to non-autonomous multi-frequency dynamics. This advancement was achieved through the Generalized Harmonic Balance (GHB) method. The core of our approach is the identification of explicit governing equations. Once these equations are learned, they provide strong physical extrapolation capabilities. This analytical representation allows the model to predict complex nonlinear responses far beyond the training point. We can then use standard numerical tools to perform rapid parametric sweeps. This strategy effectively bridges the gap between limited data observations and global nonlinear response prediction, addressing the major challenge of data acquisition in real-world MEMS applications.

We applied the proposed MEv-SINDy framework considering two high-dimensional MEMS applications. These cases included a beam resonator and a MEMS micromirror. In both examples, our method achieved highly accurate one-shot learning by predicting complex nonlinear responses from only a single training point. To ensure the reliability of the approach, we performed extensive robustness tests. We investigated the influence of training point selection on the identification accuracy. This analysis helped us define effective strategies for choosing optimal data points. Furthermore, we extended the framework to include multi-point training thanks to a frequency normalization preprocessing. The results demonstrate the use of multiple training points can refine the MEv-SINDy outcomes across different excitation levels and frequency ranges. These findings confirm the practical utility of our method for real-world engineering design. The framework significantly reduces the computational burden while preserving the complex physics of forced oscillators.

\vspace{8mm}
\textbf{Acknowledgments}: 

L.R. is supported by the Joint Research Platform “Sensor sysTEms and Advanced Materials” (STEAM) between Politecnico di Milano and STMicroelectronics. T.M, W.C., and L.Z. are supported by the National Key Research and Development Program of China (grant no.~2022YFC3005301), the National Natural Science Foundation of China (grant no.~52478552 and no.~52378527), the Natural Science Foundation of Shanghai (grant no. 23ZR1464900), the Fundamental Research Funds for the Central Universities (grant no.~22120240363) and China Scholarship Council (grant no.~20230626015)

\textbf{Contributions}: 
Conceptualization: T.M., L.R., A.F;
Methodology: T.M.;
Experiment: T.M., L.R., A.F.;
Visualization: T.M.;
Funding acquisition: W.C., L.Z.;
Results Discussion: T.M., L.R, W.C., L.Z., A.F.
Project administration: A,F.;
Supervision: A.F.;
Writing – original draft: T.M.;
Writing – review \& editing: T.M., L.R., A.F.

\textbf{Competing interests}: Authors declare that they have no competing interests.

\textbf{Code availability}: All data are provided in the main text or the supplementary materials. The codes are available in the public GitHub repository: \url{https://github.com/TengMa25/MEv-SINDy.git}

\bibliographystyle{elsarticle-harv} 
\bibliography{ref}






\end{document}